\documentclass{article}

    \PassOptionsToPackage{numbers, compress}{natbib}

 \usepackage[main, final]{neurips_2026}

\usepackage[utf8]{inputenc} 
\usepackage[T1]{fontenc}    
\usepackage{url}            
\usepackage{booktabs}       
\usepackage{amsfonts}       
\usepackage{nicefrac}       
\usepackage{microtype}      
\usepackage[table]{xcolor} 
\usepackage{graphicx}
\usepackage{amsmath}
\usepackage{mathtools}
\usepackage{enumitem} 
\usepackage{subcaption}
\usepackage[normalem]{ulem}

\usepackage{xcolor}

\usepackage[pagebackref,breaklinks,colorlinks]{hyperref}
\hypersetup{
    colorlinks=true, 
    linkcolor=blue,  
    filecolor=magenta, 
    urlcolor=blue,    
    citecolor=blue,       
}

\newcommand{\appref}[1]{\hyperref[#1]{Appendix~\ref*{#1}}}

\newcommand{\vect}[1]{\boldsymbol{#1}}
\newcommand{\mat}[1]{\mathbf{#1}}
\newcommand{\trans}{^\top}
\newcommand{\R}{\mathbb{R}}
\newcommand{\pd}[2]{\frac{\partial #1}{\partial #2}}
\definecolor{fgcpabrow}{RGB}{237,244,255}

\title{Factorized Gradients for Scalable \\ Highly-expressive Parametric Diffeomorphisms} 

 \author{%
  \textbf{Amit Aflalo}\textsuperscript{1} \quad
  \textbf{Eran Treister}\textsuperscript{1} \quad
  \textbf{Chaim Baskin}\textsuperscript{2} \quad
  \textbf{Oren Freifeld}\textsuperscript{1}
  \\[0.7em]
  \normalfont
  \textsuperscript{1}The Stein Faculty of Computer and Information Science,
  Ben-Gurion University of the Negev, Israel\\
  \textsuperscript{2}School of Electrical and Computer Engineering,
  Ben-Gurion University of the Negev, Israel
}

\begin{document}

\maketitle

\begin{figure}[htbp]
  \centering
  \vspace{-0.7cm}
  \includegraphics[width=\linewidth]{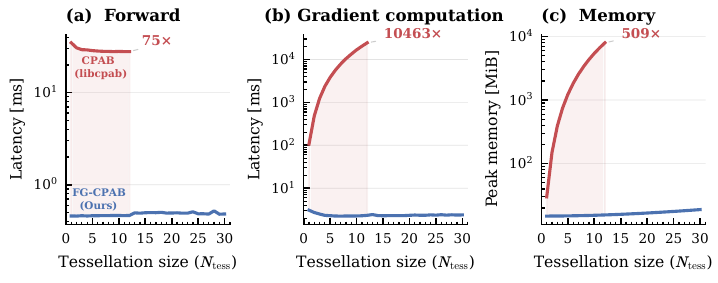}
  \vspace{-0.5cm}
  \caption{
    \textbf{Quantitative comparison, in 2D, of performance scaling with tessellation size} ($N_{\text{tess}}$),
    between the proposed method, called \textbf{FG-CPAB}, and \texttt{libcpab}~\cite{libcpab}, the standard implementation of CPAB diffeomorphisms~\cite{cpab}. The evaluation was performed 
    on a single RTX A6000 GPU with 125,000 2D points and a batch size of 16.
    \textbf{(a)} Forward pass latency. 
    \textbf{(b)} Gradient computation latency. 
    \textbf{(c)} Peak GPU memory allocation. 
    In 2D, the domain is partitioned into $C=4N_{\text{tess}}^2$ triangular cells.
     Notably, the standard implementation does not support $N_\mathrm{tess} > 12$, whereas ours scales efficiently.
  }
  \label{fig:tess_benchmark}
\end{figure}

\begin{abstract}
Diffeomorphisms  provide a flexible, topology-preserving mathematical tool for modeling complex spatial or temporal deformations, and are used in various
scientific fields.
However, making these deformations simultaneously highly expressive and computationally efficient remains challenging. Continuous Piecewise-Affine 
Based (CPAB) transformations~\cite{cpab, cpab2017}
offer an attractive solution by parameterizing a family of diffeomorphisms via continuous velocity fields that are piecewise affine \emph{w.r.t.}~a chosen tessellation of the domain. Importantly, $d=\dim(\theta)$, the dimension of
the parameter vector $\theta$,  depends on the fineness of the tessellation  rather than on the data resolution. Despite this compact parameterization, CPAB optimization has long been hindered by the tight coupling between trajectory integration and 
gradient computation \emph{w.r.t.}~$\theta$.
We introduce \textbf{FG-CPAB}, a factorized-gradient formulation for scalable CPAB transformations. By separating integration from parameter-space projection, our method reduces gradient computation complexity from \(\mathcal{O}(dTN)\) to \(\mathcal{O}(TN + d C)\), where \(T\) is the number of integration steps, \(N\) is the number of transformed points, and \(C\) is the number of tessellation cells. This reformulation yields speedups of up to four orders of magnitude (e.g., $\approx 10^4\times$ in 2D), substantial memory savings (e.g., $\approx 500\times$ reduction in peak GPU memory usage in 2D), and dramatically higher expressiveness.
This makes, for the first time, fine-tessellation CPAB practical in 2D  and 3D, enabling highly expressive parametric diffeomorphisms. 
For code and resources, see our \href{https://bgu-cs-vil.github.io/FG-CPAB/}{project page}.
\end{abstract}

\section{Introduction}
A diffeomorphism is a differentiable invertible function whose inverse is also differentiable. 
Diffeomorphisms are a natural choice in spatial modeling tasks that require smooth, invertible, and topology-preserving deformations. 
They are widely used in applications such as registration, spatial alignment, and geometric warping, where folding, tearing, or self-intersection would result in geometrically invalid deformations.
The standard way to construct such transformations is via integration of either a stationary velocity field (SVF)~\cite{svf} or a time-dependent velocity field as in Large Deformation Diffeomorphic Metric Mapping (LDDMM)~\cite{lddmm}. However, in practice, such methods often rely on a dense-grid discretization of the spatial domain. As a result, the number of parameters controlling the diffeomorphisms grows directly with the resolution of the domain (e.g., with the number of pixels in 2D). In modern 2D and especially 3D settings, this leads to high memory use, slow optimization, and difficult optimization landscapes.

Continuous Piecewise-Affine Based (CPAB) transformations, introduced by Freifeld et al.~\cite{cpab,cpab2017}, offer a compelling alternative. Instead of a dense-grid discretization, this approach tessellates the domain into cells (e.g., triangles in the 2D case) and constrains the velocity field to be a continuous piecewise-affine (CPA) function. A CPAB transformation is, by definition~\cite{cpab}, obtained by integrating a CPA velocity field. Under a mild, easily enforced regularity condition, a CPAB
transformation is a diffeomorphism\footnote{CPAB transformations are
homeomorphisms and even bi-Lipschitz functions (hence
also almost-everywhere differentiable with an almost-everywhere
differentiable inverse). 
Typically, but technically not always, they are also diffeomorphisms. For the sufficient regularity condition ensuring
that a CPAB transformation is a diffeomorphism,
see~\cite{FreifeldAflaloArxiv2026}.}. A key point is that the number of parameters controlling a CPAB transformation depends on the tessellation, not the data resolution, thereby decoupling model complexity from data resolution. These transformations provide a compact low-dimensional parameterization while still capturing rich nonlinear deformations and have been used in diverse deep learning settings, including 1D applications such as time-series alignment and trainable activation functions~\cite{Cyclic,martinez2022closed,shapira2019diffeomorphic,weber2023regularization,DiTAC,digraf, neifar2024leveraging, kryeem2023personalized, timepoint}, parameter-efficient multi-task fine-tuning~\cite{ditask}, and 2D vision and generative modeling tasks~\cite{hauberg2016dreaming,detlefsen2018deep,schwobel2022probabilistic,shacht2021single,skafte2019explicit,wang2024continuous, hartshorne2025arc}.

Despite this compact parameterization,  CPAB transformations remained hard to scale in practice. The bottleneck lies in computing the loss gradient, \(\nabla_\theta\mathcal L\), which conventionally requires propagating parameter sensitivities through the transformation \(x\mapsto\Phi^\theta(x)\). In the standard formulation, sensitivities \emph{w.r.t.}~ all $d=\dim(\theta)$ parameters are propagated through all $T$ integration steps for each of the $N$ spatial points, yielding a backward complexity of $\mathcal{O}(dTN)$. This tight coupling between parameter dimension and trajectory integration makes optimization prohibitively expensive once tessellations become fine or spatial inputs become large. Consequently, existing CPAB implementations are often limited to coarse tessellations, which restricts practical expressiveness. Moreover, even with coarse tessellations in 2D, CPAB optimization is too slow for most practical applications. In 3D, the problem worsens, and scalable CPAB optimization has remained largely out of reach, despite the wide use of diffeomorphisms in this setting~\cite{bi2026shaping,dimon,dno,meshflow,transmorph,sun2022topology,han2024hybrid,liu2025enhancing,carriere2024diffeomorphic}.

In this paper, we address this bottleneck with \textbf{FG-CPAB}, a factorized-gradient formulation that separates trajectory integration from parameter-space projection. Instead of propagating per-parameter sensitivities during integration, we perform a single adjoint sweep that accumulates low-dimensional cell-wise geometric statistics and recovers all parameter gradients through one projection onto the CPA basis. This reduces the backward complexity from $\mathcal{O}(dTN)$ to $\mathcal{O}(TN + dC)$, where $C$ is the number of tessellation cells. We further introduce a sparse basis formulation that reduces memory use and allows the final projection step to fully exploit basis sparsity, thereby realizing the full computational benefit of the factorized gradients in practice. 

Combined with optimized GPU kernels, the resulting method makes CPAB practical at scales that were previously inaccessible.
For example, in 2D, our implementation achieves up to $75\times$ faster forward passes, $10^4\times$ faster backward passes, and over $500\times$ lower peak memory compared with the standard implementation~\cite{libcpab}. Such gains enable efficient optimization of fine-tessellation CPAB diffeomorphisms
in both 2D and 3D, and make CPAB competitive as a modern primitive for large-scale nonrigid spatial modeling. To summarize, our main contributions are:
\begin{itemize}[leftmargin=*]
    \item \textbf{FG-CPAB.}
    We introduce a gradient-factorized backward pass based on discrete-adjoint moment accumulation. Unlike standard adjoint methods, which still carry parameter dependence through the backward integration, our method exploits the CPA basis structure to accumulate parameter-independent cell-wise moments and recover gradients only in a final sparse projection step.

    \item \textbf{An efficient implementation.}
    We combine factorized gradients, optimized GPU kernels, and a sparse basis construction to achieve large runtime and memory gains in practice.

    \item \textbf{Scalable diffeomorphic optimization in 2D and 3D.}
    We demonstrate that the resulting formulation makes fine-tessellation CPAB practical for spatial modeling and competitive with other diffeomorphic primitives in both 2D and 3D.
\end{itemize}

\section{Background and Related Work}
\label{sec:background}

Let $\Omega \subseteq \mathbb{R}^K$ denote the spatial domain where $K \in \{1,2,3\}$.
A diffeomorphism is a differentiable bijection
$\Phi:\Omega \to \Phi(\Omega)\subseteq \mathbb{R}^K$
with a differentiable inverse.
Such transformations preserve topology, avoiding tearing,
folding, and self-intersection.
 A standard way to construct such a diffeomorphism \(\Phi\) is to integrate a 
 well-behaved velocity field \(\vect{v}\) over time. Given an initial point \(\vect{p}^{(0)}=\vect{p}_0\in\Omega\), its trajectory follows the ordinary differential equation (ODE)
\begin{align}
\dot{\vect{p}}^{(t)} = \vect{v}(\vect{p}^{(t)},t),\quad t\in[0,1],
\label{eq:background_ode}
\end{align}
with the resulting mapping defined as $\Phi(\vect{p}_0)=\vect{p}^{(1)}$. Some frameworks, such as Large Deformation Diffeomorphic Metric Mapping (LDDMM)~\cite{lddmm}, use time-varying velocity fields. A widely used alternative is the Stationary Velocity Field (SVF) formulation~\cite{svf}, which assumes the velocity field is constant over time (i.e., $\vect{v}(\vect{p},t)=\vect{v}(\vect{p})$). This simplification significantly reduces computational complexity while still producing a rich class of diffeomorphisms.

\paragraph{Dense vs. Compact Parameterizations.}
Traditional methods for constructing diffeomorphisms~\cite{svf,lddmm} parameterize $\vect{v}$ densely on the spatial grid. While this enables highly expressive deformations, it couples the number of parameters directly to the resolution of the domain. Even in 2D, this can already involve hundreds of thousands of parameters, and in 3D, the parameter count can reach millions, leading to high memory requirements and a large parameter space that can be difficult to optimize without explicit regularization. While alternative formulations restrict dense velocity fields to finite-dimensional Lie algebras~\cite{zhang2015finite, zhang2019fast}, they do not directly address the scaling bottlenecks.

\textbf{Continuous Piecewise-Affine Based (CPAB) transformations~\cite{cpab,cpab2017}}
address this limitation by imposing structure on the velocity field.
Following~\cite{cpab}, we consider tessellations of $\Omega$ into
$C$ convex polyhedral cells, possibly unbounded, and refer to these
as \emph{convex polyhedral tessellations}.
A continuous piecewise-affine (CPA) function, \emph{w.r.t.}~a given tessellation, is a continuous function whose restriction to each cell is affine.
A \emph{simplicial tessellation} is the special case in which all cells
are simplices: intervals in 1D, triangles in 2D, and tetrahedra in 3D.
Our implementation uses simplicial tessellations when zero boundary
conditions are imposed. Otherwise, we extend $\Omega$ to $\mathbb{R}^K$
and augment the simplicial tessellation with unbounded convex polyhedral
cells, following~\cite{cpab2017}.
For simplicity, the presentation below focuses on simplicial cells,
but the formulation also applies to the additional unbounded cells.

CPA functions are Lipschitz continuous, and a CPAB transformation is
defined by integrating a CPA velocity field.
These fields form a finite-dimensional linear space, constructed as follows.
Within each cell $c$, the stationary velocity field $\vect{v}$ takes the form
\begin{align}
\vect{v}_c(\vect{p}) = \mat{A}_c \widetilde{\vect{p}},
\qquad
\vect{p}\in \text{cell } c,
\label{eq:background_affine}
\end{align}
where 
$\mat{A}_c\in\mathbb{R}^{K\times(K+1)}$ contains the affine velocity
coefficients for cell $c$, and
$\widetilde{\vect{p}}=(\vect{p}^{\top},1)^{\top}$ denotes homogeneous
coordinates. Defining the augmented matrix
$\widetilde{\mat{A}}_c=
\bigl[\begin{smallmatrix}\mat{A}_c^{\top} & \vect{0}\end{smallmatrix}\bigr]^{\top}
\in\mathbb{R}^{(K+1)\times(K+1)}$,
we can specialize~\autoref{eq:background_ode} to the stationary affine
velocity field within cell $c$, obtaining
$\dot{\widetilde{\vect{p}}}^{(t)}
=\widetilde{\mat{A}}_c\widetilde{\vect{p}}^{(t)}$.
Its closed-form solution is
$\widetilde{\vect{p}}^{(t)}
=\exp\!\bigl(t\widetilde{\mat{A}}_c\bigr)\widetilde{\vect{p}}^{(0)}$,
where $\exp$ denotes the matrix exponential.
This solution is valid as long as the trajectory remains in $c$
throughout $[0,t]$.

To ensure that the global velocity field $\vect{v}$ is continuous,
we impose linear constraints requiring the affine velocity fields
of neighboring cells to agree on their shared boundaries.
Concatenating the cell-wise affine coefficients of a CPA velocity field yields a vector
of length $D=CK(K+1)$, which must lie in the null space of the
continuity constraint system.
Let $d\le D$ denote the dimension of this null space and let
$\mat{B}\in\mathbb{R}^{D\times d}$ have columns forming a basis for it.
The stacked coefficients of any CPA velocity field can therefore
be written as

\begin{align}
\operatorname{vec}\!\bigl(\{\mat{A}_c\}_{c=1}^C\bigr) = \mat{B}\vect{\theta}
\label{eq:background_param}
\end{align}
where \(\vect{\theta}\in\R^d\) is 
the vector of parameters controlling the velocity field. Crucially, this guarantees that any choice of $\vect{\theta}$ produces a CPA velocity field, shifting the dimensionality of the optimization problem from the spatial resolution of the data to the complexity of the tessellation.  
The dimension \(d\), which controls the expressiveness of the CPA velocity fields, increases with tessellation refinement.

We remark that, as shown in~\cite{cpab}, additional optional linear constraints
such as zero boundary conditions
or ones that imply volume preservation 
can be simply added to the constraint system above. 
For a more detailed
explanation of CPAB transformations, see~\cite{cpab,cpab2017,FreifeldAflaloArxiv2026}. 

\textbf{Previous CPAB implementations.}
The original CPAB framework~\cite{cpab} and its main successor \texttt{libcpab}~\cite{libcpab} compute gradients by propagating sensitivities through the integration trajectory (see~\cite{freifeld2018deriving} for details).
This couples the spatial integration to the parameter dimension $d$, yielding a prohibitive $\mathcal{O}(dTN)$ backward complexity that restricts prior methods to coarse tessellations. Although early CPAB work~\cite{cpab,cpab2017} also considered MCMC-based inference, 
this approach scales poorly in the fine-tessellation regime and cannot provide the gradients required for end-to-end learning; see \appref{app:mcmc_comparison}.

\textbf{Adjoint methods and factorization.}
Adjoint sensitivity methods are standard for differentiating through ODE solvers with reduced memory cost~\cite{chen2018neural}. 
However, for CPAB transformations, a generic adjoint formulation does not address the main computational bottleneck: gradients still involve the dependence of the velocity field on all $d$ basis parameters. 
Moreover, as noted by ANODE~\cite{gholami2019anode}, continuous adjoint formulations can yield inconsistent gradients under forward/backward discretization mismatch. 
Our approach is therefore closer in spirit to a discretize-then-optimize adjoint, but specialized to the CPAB structure: the linear CPA basis allows trajectory-dependent terms to be accumulated as cell-local quantities before a final projection onto the parameter basis.

\section{Method}
\label{sec:method}
We introduce \textbf{FG-CPAB}, a gradient-factorization technique based on discrete-adjoint moment accumulation that decouples spatial trajectory integration from the parameter dimension. Existing CPAB implementations compute gradients \emph{w.r.t.}~the transformation parameters \(\vect{\theta}\) by propagating forward sensitivities through the trajectory, incurring complexity \(\mathcal{O}(dTN)\), where \(T\) is the number of integration steps, \(N\) is the number of spatial points, and \(d=\dim(\vect{\theta})\). Since \(d\) grows with tessellation resolution and spatial dimension, this becomes prohibitive for fine tessellations and higher-dimensional inputs.

Our method's novelty is not merely in the use of a discrete adjoint, which alone does not remove the coupling to the parameter dimension. Rather, we also exploit the linear CPA basis to reorganize adjoint contributions into cell-local geometric moments that are independent of \(d\). The full parameter gradient is then recovered by a single projection onto the basis, shifting all dependence on \(\vect{\theta}\) to this final step and reducing the backward complexity from \(\mathcal{O}(dTN)\) to \(\mathcal{O}(TN + dC)\).
\textbf{This enables a much higher $d$ (i.e., finer tessellations), hence also much more expressive
diffeomorphisms.}

Furthermore, to maximize these efficiency gains,  we also introduce a sparse formulation of the CPA basis matrix \(\mat{B}\) (detailed in \autoref{app:nodal_basis}). This sparse construction improves the practical efficiency and memory footprint of both the forward and backward passes.

\subsection{Trajectory integration and gradient computation}
In the forward pass, the affine ODE within cell \(c\) admits the update
\begin{align}
\widetilde{\vect{p}}^{(t+1)}
=\exp(h\widetilde{\mat{A}}_c)\widetilde{\vect{p}}^{(t)}
\end{align}
where \(t\) indexes integration steps of size \(h\). The update is exact provided that the trajectory remains within the same cell. Accordingly, we precompute \(\mat{T}_c=\exp(h\widetilde{\mat{A}}_c)\), so that each forward step reduces to a matrix-vector multiplication. When a step crosses a cell boundary, in both the forward and gradient computation steps, we assume that the chosen integration step $h$ is sufficiently small, so the update is well approximated by averaging the velocity over the step. This is consistent with the standard CPAB formulation~\cite{libcpab}. For the gradient computation, however, we use a different discretization. Specifically, we adopt an explicit midpoint rule (second-order Runge--Kutta),
\begin{align}\label{eq:midpoint}
  \vect{p}^{(t+1)} = \vect{p}^{(t)} + h\,\mat{A}_c\,\widetilde{\vect{p}}_{\mathrm{mid}},
  \qquad
  \vect{p}_{\mathrm{mid}} = \vect{p}^{(t)} + \tfrac{h}{2}\,\mat{A}_c\,\widetilde{\vect{p}}^{(t)}.
\end{align}
Using different forward and backward schemes is also the approach adopted in the original CPAB formulation~\cite{cpab} and is also natural from the Optimize-then-Discretize viewpoint~\cite{chen2018neural}. However, since our velocity field is usually not differentiable at inter-cell boundaries,  an Optimize-then-Discretize adjoint may yield inconsistent gradients (see~ANODE~\cite{gholami2019anode}), because the backward adjoint trajectory can pass through a different sequence of cells than the forward trajectory. We therefore adopt a Discretize-then-Optimize treatment for the backward pass, and combine it with checkpointing (see~\autoref{subsec:checkpointing}) for memory efficiency. See \appref{app:integrator_mismatch} for discussion and a numerical gradient check.

\subsection{Adjoint factorization via moment accumulation}
\label{subsec:moments}
Let \(\mat{A}_{c,\mathrm{lin}} \in \mathbb{R}^{K\times K}\) denote the linear block of \(\mat{A}_c\), and 
let \(\mat{B}_{i,c}\) denote the \(i\)-th basis field's affine matrix in cell \(c\).
Differentiating \autoref{eq:midpoint} \emph{w.r.t.}~\(\theta_i\) (the $i$-th component of $\vect{\theta}$), using \(\partial \mat{A}_c/\partial \theta_i = \mat{B}_{i,c}\), yields the sensitivity recursion
\begin{align}
\label{eq:sensitivity}
\pd{\vect{p}^{(t+1)}}{\theta_i}
=
h\,\mat{B}_{i,c}\,\widetilde{\vect{p}}_{\mathrm{mid}}^{(t)}
+
\tfrac{h^2}{2}\,\mat{A}_{c,\mathrm{lin}}\,\mat{B}_{i,c}\,\widetilde{\vect{p}}^{(t)}
+
\mat{J}\,\pd{\vect{p}^{(t)}}{\theta_i},
\end{align}
\begin{align}
\text{where}\quad
\mat{J}
\coloneqq
\frac{\partial\vect{p}^{(t+1)}}{\partial\vect{p}^{(t)}}
=
\mat{I}
+
h\,\mat{A}_{c,\mathrm{lin}}
+
\tfrac{h^2}{2}\mat{A}_{c,\mathrm{lin}}^2
\end{align}
is the Jacobian of one midpoint step with respect to the current position.%
Note that \(\mat{J}\) depends on  \(\mat{A}_{c,\mathrm{lin}}\) but not on the translation component of \(\mat{A}_c\)  (see \autoref{app:jacobian_translation}). 
The translation still influences the forward trajectory through the positions themselves.

A direct implementation of \autoref{eq:sensitivity} would still require storing or propagating the full \(N \times d\) sensitivities. 
Instead, we introduce the adjoint vector \(\vect{\lambda}^{(t)}\), defined as the gradient of the loss \emph{w.r.t.}~the position at time \(t\), and propagate it backward in time. The adjoint satisfies the discrete recurrence
\begin{align}
\label{eq:adjoint}
\vect{\lambda}^{(t)}
=
\bigl(
\mat{I}
+
h\,\mat{A}_{c,\mathrm{lin}}^\top
+
\tfrac{h^2}{2}(\mat{A}_{c,\mathrm{lin}}^\top)^2
\bigr)\,\vect{\lambda}^{(t+1)}
=
\mat{J}^\top \vect{\lambda}^{(t+1)}.
\end{align}
Using this adjoint recurrence and left-multiplying \autoref{eq:sensitivity} by \(\vect{\lambda}^{(t+1)\top}\) yields
\begin{align}
\label{eq:explicit_adjoint_sensitivity}
\vect{\lambda}^{(t+1)\top}\pd{\vect{p}^{(t+1)}}{\theta_i}
=
\vect{\lambda}^{(t+1)\top}
\Bigl(
h\,\mat{B}_{i,c}\,\widetilde{\vect{p}}_{\mathrm{mid}}^{(t)}
+
\tfrac{h^2}{2}\,\mat{A}_{c,\mathrm{lin}}\,\mat{B}_{i,c}\,\widetilde{\vect{p}}^{(t)}
\Bigr)
+
\vect{\lambda}^{(t)\top}\pd{\vect{p}^{(t)}}{\theta_i}.
\end{align}
The sensitivity recurrence propagates basis contributions through
subsequent timesteps, making it difficult to group them by cell
and factor out the basis.
We note, however, that summing over all timesteps produces a telescoping series, so that
all propagated terms cancel exactly (see \autoref{app:telescoping}),
leaving local contributions that can be grouped by cell to factor
out the basis. Concretely, let $\mathcal{L}$ denote a loss and let 
\(c_n^{(t)}\) denote the cell containing point \(n\) at step \(t\). Then
\begin{align}
\label{eq:theta_local_sum}
\pd{\mathcal{L}}{\theta_i}
&=
\sum_{n=1}^{N}\sum_{t=0}^{T-1}
\vect{\lambda}_n^{(t+1)\top}
\Bigl(
h\,\mat{B}_{i,c_n^{(t)}}\,
\widetilde{\vect{p}}_{n,\mathrm{mid}}^{(t)}
+
\tfrac{h^2}{2}\,
\mat{A}_{c_n^{(t)},\mathrm{lin}}\,
\mat{B}_{i,c_n^{(t)}}\,
\widetilde{\vect{p}}_n^{(t)}
\Bigr).
\end{align}

\paragraph{Moment accumulation.}
The remaining direct terms are bilinear in the adjoint
\(\vect{\lambda}\) and the homogeneous position
\(\widetilde{\vect{p}}\).
Since the basis slice \(\mat{B}_{i,c}\) is fixed for each cell \(c\),
we use the Frobenius identity
$
\vect{\lambda}^{\top}\mat{A}\mat{B}_{i,c}\widetilde{\vect{p}}
=
\langle \mat{B}_{i,c},
(\mat{A}^{\top}\vect{\lambda})\widetilde{\vect{p}}^{\top}
\rangle_F
$
to separate the basis from the trajectory-dependent factors.
We then accumulate these factors over all visits to each cell.
Defining
\(\mathcal{I}_c \coloneqq \{(n,t)\mid c_n^{(t)}=c\}\),
the moment matrix for cell \(c\) is
\begin{align}
\label{eq:moment}
\mat{M}_c
=
\sum_{(n,t)\in \mathcal{I}_c}
\Bigl[
h\,\vect{\lambda}_n^{(t+1)}\,
\widetilde{\vect{p}}_{n,\mathrm{mid}}^{(t)\top}
+
\tfrac{h^2}{2}\,
(\mat{A}_{c,\mathrm{lin}}^\top\vect{\lambda}_n^{(t+1)})\,
\widetilde{\vect{p}}_n^{(t)\top}
\Bigr].
\end{align}
Crucially, \(\mat{M}_c \in \mathbb{R}^{K \times (K+1)}\) is decoupled from $\mat{B}_{i,c}$, hence independent of the  dimension $d$.

\paragraph{Gradient recovery.}
\label{subsec:gradient_recovery}
Once the moment matrices have been accumulated, all parameter gradients are recovered by a single sparse basis projection
\begin{align}
\label{eq:grad}
\boxed{
\pd{\mathcal{L}}{\theta_i}
=
\sum_{c=1}^{C}
\langle \mat{B}_{i,c},\, \mat{M}_c \rangle_F,
\qquad
\text{equivalently}
\qquad
\nabla_{\!\vect{\theta}}\mathcal{L}
=
\mat{B}^\top \operatorname{vec}\!\bigl(\{\mat{M}_c\}_{c=1}^C\bigr)}~.
\end{align}
The core factorization result is~\autoref{eq:grad}. 
The trajectory-dependent computation is independent of \(d\), with all explicit \(d\)-dependence confined to the final projection step. In practice, this final projection further benefits from a sparse basis construction of \(\mat{B}\), which we derive in \autoref{app:nodal_basis}. This sparse formulation reduces the memory footprint of the basis representation and improves the efficiency of the projection step.

\subsection{Memory-efficient adjoint via segmented checkpointing}
\label{subsec:checkpointing}
While factorization removes dependence on \(d\), the reverse sweep still requires access to intermediate forward states. In our GPU implementation, each thread processes a single spatial point, so storing the full trajectory would require \(\mathcal{O}(TN)\) storage across all threads (equivalently, \(\mathcal{O}(T)\) per thread), which increases register pressure and can spill to local memory.
Thus, we use segmented checkpointing with segment length \(S\) (we use $S=4$). Instead of storing all \(T+1\) states along the trajectory, each thread stores only the states at segment boundaries, giving \(\lceil T/S\rceil\) checkpoints per point. During the reverse sweep, each segment is replayed from its checkpoint and immediately used for adjoint propagation and moment accumulation.
This reduces storage from \(\mathcal{O}(TN)\) to \(\mathcal{O}(N(T/S + S))\) overall, or from \(\mathcal{O}(T)\) to \(\mathcal{O}(T/S + S)\) per thread, at the cost of only constant-factor recomputation. See \autoref{app:segmented_checkpointing}.

\subsection{Gradient computation algorithm}
\label{subsec:algorithm}

Incorporating segmented checkpointing, the factorization yields three sequential phases:
\begin{enumerate}[leftmargin=*,itemsep=2pt]
\item \textbf{Forward checkpointing} ($\mathcal{O}(TN)$).\;
  Integrate the trajectory forward, caching only sparse checkpoints at boundaries of length $S$.
\item \textbf{Adjoint sweep + moment accumulation} ($\mathcal{O}(TN)$).\;
  Iterate backward through the segments. For each segment, recompute the local forward trajectory from the checkpoint, propagate the adjoint via the transposed midpoint rule (\autoref{eq:adjoint}), and atomically accumulate the products into $\mat{M}_c$ (\autoref{eq:moment}).
\item \textbf{Sparse basis projection} ($\mathcal{O}(d\,C)$).\;
  Recover all $d$ gradients via \autoref{eq:grad}.
\end{enumerate}
The total gradient computation cost is $\mathcal{O}(TN + d\,C)$: integration and adjoint propagation are independent of $d$, and projection is independent of $N$ and $T$.

\section{Results}

In this section, we evaluate \textbf{FG-CPAB} along several axes: computational efficiency vs.~the standard CPAB implementation, integration runtime vs.~optimized diffeomorphic baselines, expressiveness and optimization behavior vs.~dense SVFs, performance on 3D deformable registration, and CPAB as a low-dimensional geometry code for operator learning. We also show that 
composing CPAB transformations 
(which are not algebraically closed under composition)
increases expressiveness further, and conclude with ablations on tessellation resolution, basis sparsity, and solver step count.

\subsection{Acceleration and memory savings over the standard CPAB implementation}
\label{subsec:acceleration_baseline}

FG-CPAB substantially improves CPAB scaling over the standard \texttt{libcpab}~\cite{libcpab} implementation. As  \autoref{fig:tess_benchmark} shows, the gains are largest in the gradient computation phase, where standard CPAB is bottlenecked by parameter-dependent sensitivity propagation.

In 2D, FG-CPAB achieves up to $75\times$ faster forward passes, $10^4\times$ faster gradient computation, and over $500\times$ lower peak memory. Under our benchmark setup, the standard implementation does not run beyond \(N_{\mathrm{tess}}=12\) because its backward pass couples the parameter dimension directly to kernel-launch geometry (see \appref{app:launch_limit}), whereas FG-CPAB continues to scale to finer resolutions. This pattern is similar in other dimensions (see \autoref{sec:appendix_performance}): in 3D, up to $60\times$ forward acceleration, $4{,}477\times$ backward acceleration, and a $180\times$ reduction in peak memory; in 1D, up to a $301\times$ backward speedup. These results match the proposed complexity reduction: by decoupling trajectory integration from parameter-space projection, FG-CPAB removes the main bottleneck that has historically restricted CPAB to coarse tessellations.

\subsection{Forward and gradient computation runtime}

We next evaluate FG-CPAB against optimized VoxelMorph SVF~\cite{voxelmorph} and \texttt{torch-lddmm}~\cite{lee_torch_lddmm} baselines on standard 2D and 3D grids. All methods are run with a batch size of 32. FG-CPAB uses a $15 \times 15$ tessellation in 2D ($4\times15^2$ triangles) and a $15 \times 15 \times 15$ tessellation in 3D ($5\times15^3$ tetrahedra). For each method, we measure the latency of the forward spatial transformation and the latency of computing the gradients \emph{w.r.t.}~the transformation parameters \(\vect{\theta}\).

FG-CPAB is faster than both baselines, as reported in \autoref{tab:integration_times}. In 2D, it reduces total runtime by \(2.4\times\) relative to VoxelMorph and by \(21.4\times\) relative to \texttt{torch-lddmm}. The same trend holds in 3D, with speedups of \(2.3\times\) and \(21.2\times\), respectively. These improvements are due to the combination of CPAB's low-dimensional parameterization and the proposed factorized gradient computation, which avoids propagating parameter-dependent sensitivities during trajectory integration.

\begin{table}[htpb]
  \centering
  \vspace{-0.2cm}
    \captionsetup{font=footnotesize} 

  \scriptsize
  \setlength{\tabcolsep}{3.0pt}
  \resizebox{\linewidth}{!}{%
  \begin{tabular}{lcccccc}
    \toprule
    & \multicolumn{3}{c}{\textbf{2D Integration} ($512 \times 512$)} & \multicolumn{3}{c}{\textbf{3D Integration} ($128 \times 128 \times 128$)} \\
    \cmidrule(lr){2-4} \cmidrule(lr){5-7}
    \textbf{Method} & \textbf{Forward (ms)} & \textbf{Gradient (ms)} & \textbf{Peak Mem (MiB)} & \textbf{Forward (ms)} & \textbf{Gradient (ms)} & \textbf{Peak Mem (MiB)} \\
    \midrule
    \rowcolor{fgcpabrow}
    FG-CPAB & $\mathbf{1.69 \pm 0.06}$ & $\mathbf{8.64 \pm 0.02}$ & $\mathbf{197.97}$ & $\mathbf{31.10 \pm 0.18}$ & $\mathbf{117.50 \pm 0.34}$ & $\mathbf{2351.61}$ \\
    VoxelMorph SVF & $11.41 \pm 0.02$ & $13.42 \pm 0.08$ & $1330.13$ & $119.14 \pm 0.03$ & $220.21 \pm 0.24$ & $14432.13$ \\
    torch-lddmm & $51.23 \pm 0.41$ & $169.75 \pm 0.06$ & $3162.13$ & $732.81 \pm 0.40$ & $2414.12 \pm 0.53$ & $43288.13$ \\
    \bottomrule
  \end{tabular}%
  }
  \caption{Forward transformation and parameter gradient ($\nabla_{\!\vect{\theta}}$) latency, alongside peak memory footprint, on 2D and 3D data.}
  \label{tab:integration_times}
  \vspace{-0.7cm}
\end{table}

\subsection{Expressiveness and optimization stability compared with dense SVFs}
\label{sec:cpab_vs_dense}

We compare FG-CPAB against a dense stationary velocity field (SVF) baseline on a synthetic registration benchmark designed to isolate optimization behavior under a matched budget (1000 fit steps). In both 2D ($512 \times 512$) and 3D ($128 \times 128 \times 128$), we generate blob-based source images, warp them with sampled smooth deformations to form targets, and fit either a CPAB transformation (using FG-CPAB) or a dense SVF integrated with 7 scaling-and-squaring steps. FG-CPAB uses a $15 \times 15$ tessellation in 2D ($4\times15^2$ triangles) and a $15 \times 15 \times 15$ tessellation in 3D ($5\times15^3$ tetrahedra); full implementation details are provided in \autoref{app:impl_details}.

\begin{figure}[htbp]
  \centering
  \includegraphics[width=\linewidth]{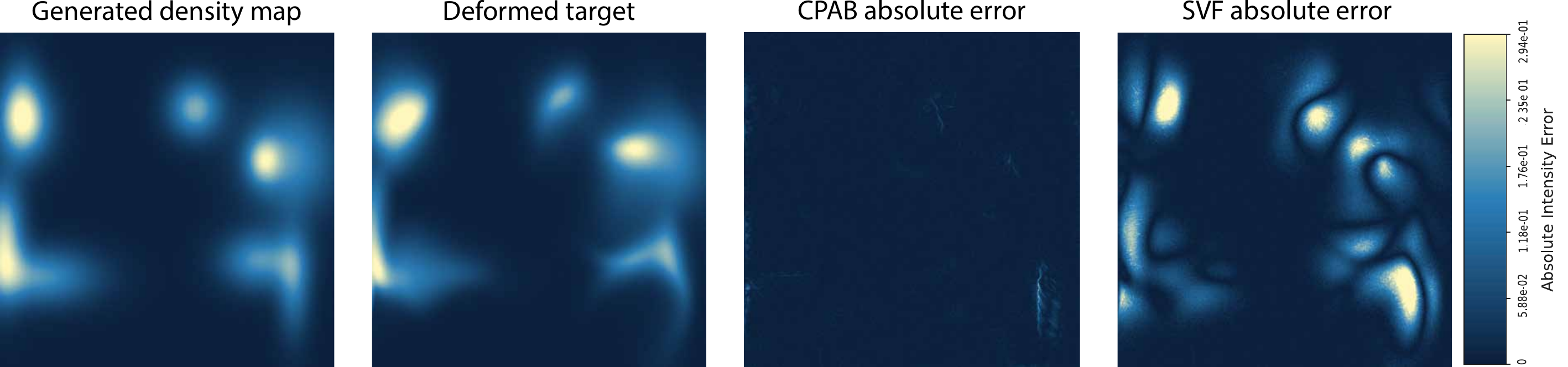}
  \vspace{-0.2cm}
\caption{
    \textbf{Qualitative comparison of CPAB and dense SVF on synthetic deformation fitting.}
    Left to right: Source image, deformed target, absolute alignment error after optimization with CPAB, and absolute alignment error after optimization with dense SVF under the same optimization budget.
    CPAB achieves near-zero residual error despite its compact parameterization, whereas dense SVF more often exhibits structured residual artifacts under the matched budget.
}
  \label{fig:cpab_vs_svf}
\end{figure}

\autoref{tab:svf_vs_cpab} reports reconstruction error, fitting time, and the fraction of non-positive Jacobian determinants, while \autoref{fig:cpab_vs_svf} shows  qualitative results. Under the same optimization budget, FG-CPAB consistently reaches lower residual error, converges faster, and exhibits fewer topology violations. These gains arise from the compact CPAB parameterization: optimizing within the low-dimensional space spanned by $\mat{B}$ acts as an implicit regularizer, substantially improving optimization behavior, potentially due to a smoother optimization landscape than that of a dense 
SVF parameterization.

\begin{table}[!b]
  \vspace{-0.4cm}
  \centering
  \captionsetup{font=footnotesize} 
  
  \scriptsize
  \setlength{\tabcolsep}{3.0pt}
  \resizebox{\linewidth}{!}{%
    \begin{tabular}{lcccccc}
      \toprule
      Dim & Params (C,S) & MSE C $\downarrow$ & MSE S $\downarrow$ & Fit C (s) $\downarrow$ & Fit S (s) $\downarrow$ & $|J|\leq0$ (\%) (C,S) $\downarrow$ \\
      \midrule
      2D & 842 / 524288 & \textbf{8.853e-05 $\pm$ 3.9e-05} & 2.651e-03 $\pm$ 1.2e-03 & \textbf{0.525 $\pm$ 0.044} & 1.356 $\pm$ 0.007 & \textbf{0.028\%} / 0.178\% \\
      3D & 8232 / 6291456 & \textbf{7.255e-05 $\pm$ 2.5e-05} & 1.676e-04 $\pm$ 4.1e-05 & \textbf{6.140 $\pm$ 0.021} & 12.906 $\pm$ 0.021 & \textbf{0.012\%} / 0.021\% \\
      \bottomrule
    \end{tabular}%
  }
  \caption{Quantitative synthetic registration results comparing FG-CPAB (C) and dense SVF (S) under the same optimization budget in 2D and 3D. Metrics report final reconstruction MSE, wall-clock fitting time, and the percentage of spatial locations with non-positive Jacobian determinant.}
  \label{tab:svf_vs_cpab}
  \vspace{-0.25cm}
\end{table}

Importantly, our claim is not that dense SVFs lack representational power. Rather, the comparison highlights a practical trade-off between asymptotic flexibility and optimization efficiency. Dense SVFs can represent highly detailed deformations, but their large parameter space makes them much harder to optimize reliably under realistic compute budgets. In contrast, CPAB offers a more compact parameterization that appears to facilitate efficient optimization while retaining sufficient expressiveness for the evaluated deformations.

To further probe this trade-off, \appref{app:impl_synthetic_cpab_svf} studies extended-budget optimization and regularized dense SVFs. Those results show that heavily regularized dense SVFs can eventually approach, and in some cases surpass, CPAB in final reconstruction error, but only after orders of magnitude more optimization steps. Thus, the main advantage of CPAB is practical: it achieves strong alignment quality much more reliably and with substantially less compute.

\subsection{CPAB for 3D deformable registration}

We evaluate FG-CPAB in the classical \emph{instance-specific optimization} setting for deformable registration. For each test volume pair, we directly optimize the compact CPAB parameter vector \(\vect{\theta}\) (using the factorized gradients from \autoref{sec:method}). No offline pre-training on any dataset is performed. This setup highlights the practical ease of optimization unlocked by our method: the compact parameterization and \(\mathcal{O}(TN + d C)\) backward pass enable fast, stable, topology-preserving convergence per volume pair. Full implementation details, including the optimizer and loss function, are provided in \appref{app:impl_ixi_cpab_registration}.

\autoref{tab:registratioComparison} reports results on the IXI brain MRI benchmark. A single CPAB transformation reaches \(0.760 \pm 0.023\) Dice with near-perfect topology preservation (\(<0.0001\%\) non-positive Jacobians). This performance is competitive with strong learned baselines, despite requiring no offline pre-training. Direct per-pair optimization converges in \(\approx 5\) seconds per volume pair on an RTX A6000 GPU, compared with \(\approx 135\) seconds for LDDMM. FG-CPAB therefore offers a substantially better accuracy-runtime trade-off than classical optimization-based diffeomorphic methods.

We further observe that sequential composition of CPAB transformations increases expressiveness while preserving topology. Composing \(i\) transformations raises the Dice score from \(0.760\) (\(i=1\)) to \(0.765\) (\(i=4\)), with the fraction of non-positive Jacobians remaining below \(0.0001\) throughout. These results demonstrate that the proposed factorized formulation makes fine-tessellation CPAB a practical, optimization-friendly primitive for 3D deformable registration.

\begin{center}
  \begin{minipage}[c]{0.50\linewidth}
    \centering
    \captionof{table}{Registration performance comparison on the IXI dataset. For FG-CPAB, $i$ indicates the number of composed transformations.}
    \label{tab:registratioComparison}
    \setlength{\tabcolsep}{5pt}
    \resizebox{0.975\linewidth}{!}{%
    \begin{tabular}{lcc}
      \toprule
      \textbf{Model} & \textbf{Dice} & \textbf{ $|J| \le 0$ (\%)} \\
      \midrule
      Affine & $0.386 \pm 0.195$ & - \\
      SyN & $0.645 \pm 0.152$ & $< 0.0001$ \\
      NiftyReg & $0.645 \pm 0.167$ & $0.020 \pm 0.046$ \\
      LDDMM & $0.680 \pm 0.135$ & $< 0.0001$ \\
      VoxelMorph & $0.732 \pm 0.123$ & $1.522 \pm 0.336$ \\
      CycleMorph & $0.737 \pm 0.123$ & $1.719 \pm 0.382$ \\
      MIDIR & $0.742 \pm 0.128$ & $< 0.0001$ \\
      ViT-V-Net & $0.734 \pm 0.124$ & $1.609 \pm 0.319$ \\
      CoTr & $0.735 \pm 0.135$ & $1.298 \pm 0.343$ \\
      PVT & $0.727 \pm 0.128$ & $1.858 \pm 0.314$ \\
      nnFormer & $0.747 \pm 0.135$ & $1.595 \pm 0.358$ \\
      TransMorph & $0.753 \pm 0.123$ & $1.579 \pm 0.328$ \\
      ScaMorph & $0.755 \pm 0.132$ & $1.400 \pm 0.329$ \\
      \midrule
      \rowcolor{fgcpabrow}
      FG-CPAB & $0.760 \pm 0.023$ & $< 0.0001$ \\
      \rowcolor{fgcpabrow}
      FG-CPAB ($i=2$) & ${0.762 \pm 0.024}$ & $< 0.0001$ \\
      \rowcolor{fgcpabrow}
      FG-CPAB ($i=3$) & ${0.763 \pm 0.024}$ & $< 0.0001$ \\
      \rowcolor{fgcpabrow}
      FG-CPAB ($i=4$) & $\mathbf{0.765 \pm 0.024}$ & $< 0.0001$ \\
      \bottomrule
    \end{tabular}%
    }
  \end{minipage}\hfill
  \begin{minipage}[c]{0.43\linewidth}
    \centering
    \captionof{figure}{Dice score vs.~tessellation size on IXI. At tessellation 30 we have $5\times30^3=135,000$ unique cells.}
    \includegraphics[width=\linewidth]{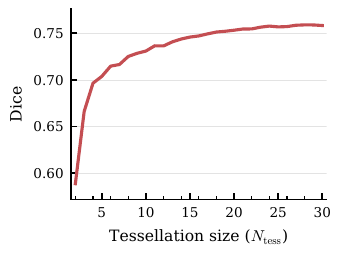}
    \label{fig:ixi_sweep}
  \end{minipage}
\end{center}

\subsection{CPAB as a low-dimensional geometry code for neural operator learning}
\label{subsec:cpab_operator}

DIMON~\cite{dimon} learns geometry-dependent PDE operators by transporting inputs and solutions from varying domains \(\{\Omega^\theta\}\) to a reference domain \(\Omega^0\), and conditioning the operator on a low-dimensional geometry code. In the original DIMON pipeline, this code is obtained by compressing LDDMM high-dimensional deformation representation with principal component analysis (PCA).

We consider the same reaction--diffusion setting, but replace the geometry encoding with CPAB. For each target domain \(\Omega^\theta\), we fit a CPAB diffeomorphism \(\Phi_\theta:\Omega^0\to\Omega^\theta\) and use the corresponding parameter vector \(\vect{\theta}\in\mathbb{R}^d\) directly as the geometry code. Unlike PCA-based LDDMM encodings, CPAB provides a native low-dimensional diffeomorphic parameterization and therefore removes the need for a separate compression step. We match the CPAB parameter count to that of the PCA encoding.

This has two advantages: the code is directly tied to the transport map, and its dimension is controlled by the tessellation, yielding an interpretable expressiveness--complexity trade-off. \autoref{tab:rd_cpab_operator} shows that FG-CPAB attains the lowest test relative \(L^2\) error under the same operator backbone, suggesting that compact diffeomorphic parameterizations can serve as effective geometry descriptors for neural operator learning. We defer implementation details to \appref{app:impl_cpab_operator}.
\begin{table}[htbp]
  \centering
  \small
  \begin{tabular}{lccc}
    \toprule
    \textbf{Experiment} & \textbf{Geometry model} & \textbf{Test rel. \(L^2\) mean} & \textbf{Test rel. \(L^2\) std} \\
    \midrule
    \rowcolor{fgcpabrow}
    FG-CPAB & FG-CPAB parameterization & \textbf{0.052811} & \textbf{0.022436} \\
    LDDMM deformation PCA & PCA deformation code & 0.082821 & 0.033168 \\
    \bottomrule
  \end{tabular}
  \vspace{0.1cm}
  \caption{Reaction--diffusion operator-learning results using different geometry encodings. Lower is better for both metrics.}
  \label{tab:rd_cpab_operator}
  \vspace{-1cm}
\end{table}

\subsection{CPAB composition for non-stationary velocity fields}
\label{subsec:cpab_composition}
\autoref{tab:registratioComparison} shows that FG-CPAB already performs competitively with, and in our setting better than, LDDMM. However, the two models differ in expressiveness: a single CPAB transformation is generated by a stationary CPA velocity field, whereas LDDMM can represent more general deformations through time-varying velocity fields. Therefore, there exist 
orientation-preserving diffeomorphisms that cannot be represented (or even well approximated) by a single CPAB diffeomorphism.
Fortunately, the efficiency of our implementation also makes it practical to compose $m$ CPAB transformations,
\begin{align}
\Phi = \Phi_m \circ \cdots \circ \Phi_1.
\end{align}

Although such a composition is not necessarily itself a CPAB transformation (CPAB diffeomorphisms form a subset, but not a subgroup, of the diffeomorphism group and are not closed under composition~\cite{cpab}), it remains a diffeomorphism, since diffeomorphisms are closed under composition. This provides a practical way to increase expressiveness while retaining CPAB's efficiency at each stage.

\begin{figure}[htbp]
  \centering
  \includegraphics[width=\linewidth]{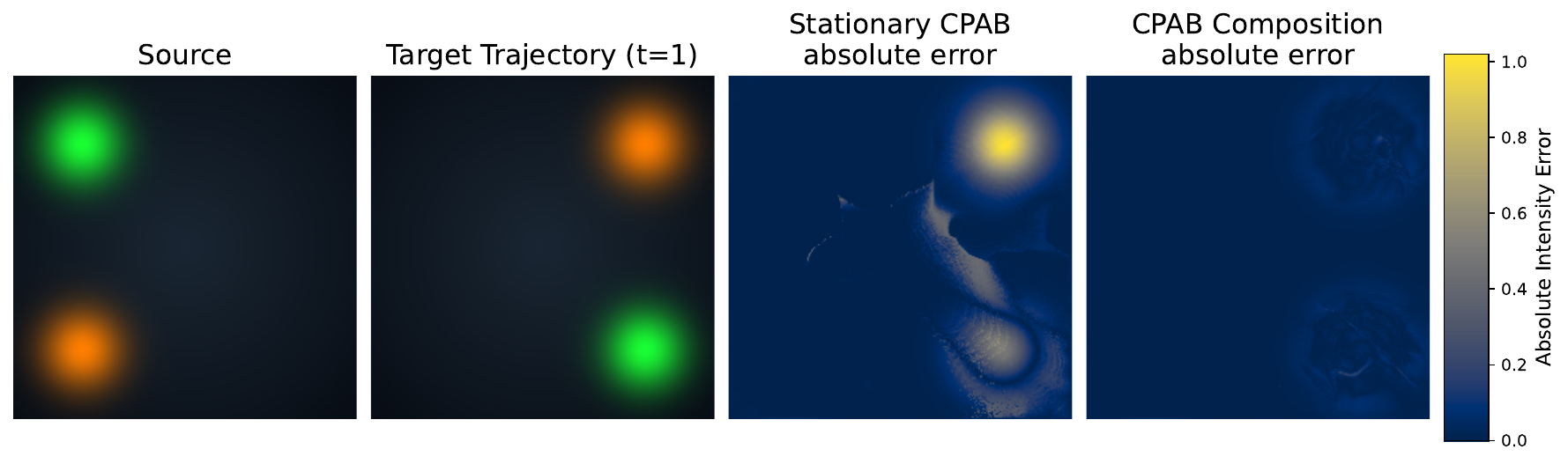}
  \vspace{-0.2cm}
\caption{
\textbf{A composition of CPAB transformations can model deformations that a single stationary CPAB cannot.}
From left to right: source image, target image, absolute error for a single CPAB transformation, and absolute error for a composition of CPAB transformations. The target requires the two blobs to follow crossing trajectories, which cannot be captured well by a single stationary velocity field. While a single CPAB leaves substantial residual error, the composed model greatly reduces the error by realizing the deformation through multiple diffeomorphic stages.
}
  \label{fig:cpab_comp}
\end{figure}

\autoref{fig:cpab_comp} illustrates this point on a synthetic example with two blobs following crossing trajectories. A single CPAB transformation cannot model this motion accurately and leaves a visible residual error. In contrast, a composition of CPAB transformations successfully captures the deformation by distributing the motion across multiple stages. Thus, CPAB composition provides a simple and efficient way to approximate deformations that would otherwise require a non-stationary velocity field, bringing CPAB closer in spirit to the flexibility of LDDMM, but at a much lower cost. 

\subsection{Ablations and hyper-parameter analysis}
\label{subsec:ablation}

\paragraph{Basis sparsity.}
We first ablate the sparse basis construction by comparing sparse and dense representations of the CPA basis. The sparse formulation yields substantial memory savings, with additional runtime benefits that become more pronounced at larger tessellation sizes. This shows that sparsity is an important complement to gradient factorization for making high-resolution CPAB practical. Full results are provided in \appref{app:sparse_abl}.

\paragraph{Tessellation resolution.}
We next study tessellation resolution, the main expressiveness knob of CPAB. Performance improves as the tessellation becomes finer, both on IXI (\autoref{fig:ixi_sweep}) and in the synthetic setting, showing that the additional local flexibility is useful in practice. This result also highlights the importance of our efficient backward pass, which makes these larger tessellations practical to optimize. Full results are provided in \appref{app:tess_res}.

\paragraph{Solver step count.}
We also study the number of ODE solver steps used for CPAB integration. All results in the paper use 50 CPAB solver steps unless noted otherwise. Accuracy improves rapidly at low step counts and then largely plateaus around 50 steps, while runtime continues to increase. This supports using 50 steps as a robust default, consistent with previous CPAB implementations. Full results are provided in \appref{app:solver_steps}.

\section{Discussion}
\label{sec:discussion}
Our results show that \textbf{FG-CPAB} makes CPAB substantially more practical as a diffeomorphic primitive for modern spatial or temporal  modeling. By decoupling trajectory integration from parameter-space projection, the proposed formulation removes the main computational bottleneck of standard CPAB and enables efficient optimization with much finer tessellations in both 2D and 3D. In turn, this improved scalability translates into better practical expressiveness, competitive registration accuracy, and strong topology preservation.

While adjoint methods are standard for general ODEs~\cite{chen2018neural}, the linear CPA basis permits an exact factorization into cell-local moments that prior general-purpose adjoint solvers cannot exploit. The key insight is the reorganization of sensitivities into parameter-independent geometric statistics that are later projected through the sparse CPA basis. This delivers the \(\mathcal{O}(TN + d C)\) complexity and the dramatic practical speedups reported here.

{These efficiency gains, however, do not eliminate the dimensionality limitations of tessellation-based models.} Although the factorization extends directly to any spatial dimension \(K\), with \(K\times(K+1)\) moments and parameter dependence confined to the final projection, the number of cells grows exponentially with \(K\). We therefore restrict our empirical claims to spatial dimensions \(K\leq3\). Extending the approach to higher-dimensional domains would likely require additional structure, such as separable, low-rank, or adaptively refined velocity fields.

Overall, this work shifts CPAB from a promising but difficult-to-scale formulation toward a useful, optimization-friendly tool for large-scale diffeomorphic learning in both classical optimization and modern deep-learning pipelines.

\newpage

\bibliographystyle{plainnat}
\bibliography{references}

\newpage
\appendix

\section{Performance}
\label{sec:appendix_performance}

In this section, we evaluate the computational efficiency of our optimized CPAB implementation against the strongest available baselines. In 2D and 3D, we compare against the standard \texttt{libcpab}~\cite{libcpab} implementation. In 1D, we additionally compare against \texttt{DIFW}~\cite{martinez2022closed}, which to our knowledge is the fastest available implementation in the 1D setting, although it only supports 1D transformations. To ensure reliable measurements, all reported results represent the mean of 50 execution iterations, preceded by a 10-iteration warmup phase. All experiments were performed on a standardized workload of 125,000 points with a batch size of 16.

\subsection{Why the standard backward stops scaling}
\label{app:launch_limit}

The failure of the standard implementation at fine tessellation is not an intrinsic limitation of CPAB, but a consequence of how the reference backward pass is organized on the GPU. In previous implementations, sensitivities are propagated separately for every parameter throughout the full integration trajectory. As tessellation becomes finer, the parameter dimension \(d\) grows, and the backward pass therefore scales directly with \(d\).

The key hardware bottleneck is that the parameter axis is encoded directly in the backward-kernel launch geometry. As a result, increasing tessellation enlarges the kernel grid itself and eventually reaches a device-dependent launch ceiling; this is the \emph{launch-limit problem} referred to in the main text. In practice, once tessellations become moderately fine, the reference implementation is governed by this parameter-dependent kernel structure rather than by the geometric integration alone.

FG-CPAB avoids this bottleneck by removing parameter dependence from the integration itself. The fused adjoint kernel is launched only over points and batch elements, accumulates parameter-independent cell-wise moments, and defers the dependence on \(d\) to a separate final projection step. This is the practical reason the optimized implementation continues to run at tessellation levels for which the standard backward becomes unusable.

\begin{figure}[htbp]
  \centering
  \includegraphics[width=\linewidth]{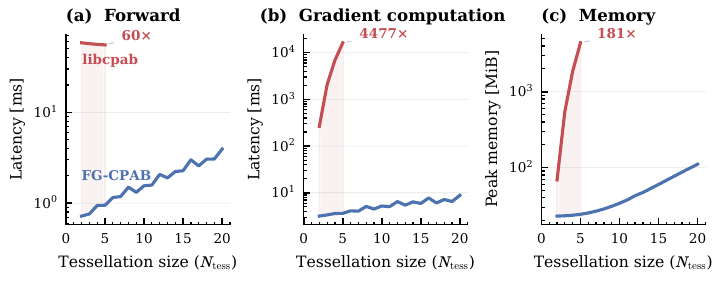}
  \vspace{-0.2cm}
  \caption{
    \textbf{Performance scaling with tessellation size ($N_\mathrm{tess}$).} 
    Comparison of our optimized 3D CPAB vs.~standard implementation on a single RTX A6000 GPU. 
    \textbf{(a)} Forward pass latency. 
    \textbf{(b)} Gradient computation latency. 
    \textbf{(c)} Peak GPU memory allocation. 
    Our approach yields up to a $60\times$ speedup in the forward pass and a $4477\times$ speedup in the gradient computation, while reducing peak memory by over $180\times$. Notably, the standard implementation does not support $N_\mathrm{tess} > 5$, whereas ours scales efficiently.
  }
  \label{fig:tess_benchmark_3d}
\end{figure}

\begin{figure}[htbp]
  \centering
  \includegraphics[width=\linewidth]{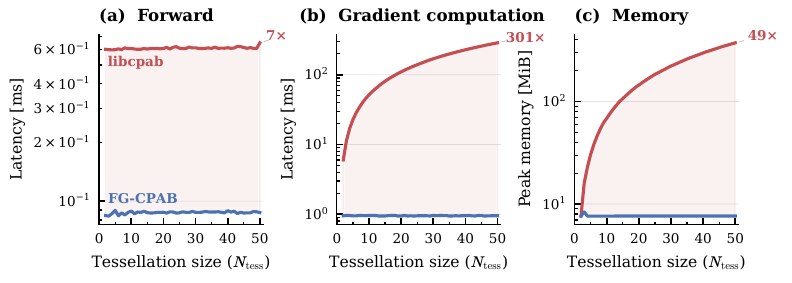}
  \vspace{-0.2cm}
  \caption{
    \textbf{Performance scaling with tessellation size ($N_\mathrm{tess}$).} 
    Comparison of our optimized 1D CPAB vs.~\texttt{libcpab} implementation on a single RTX A6000 GPU. 
    \textbf{(a)} Forward pass latency. 
    \textbf{(b)} Gradient computation latency. 
    \textbf{(c)} Peak GPU memory allocation. 
    Our approach yields up to a $7\times$ speedup in the forward pass and a $301\times$ speedup in the gradient computation, while reducing peak memory by $49\times$.
  }
  \label{fig:tess_benchmark_1d}
\end{figure}

\begin{figure}[htbp]
  \centering
  \includegraphics[width=\linewidth]{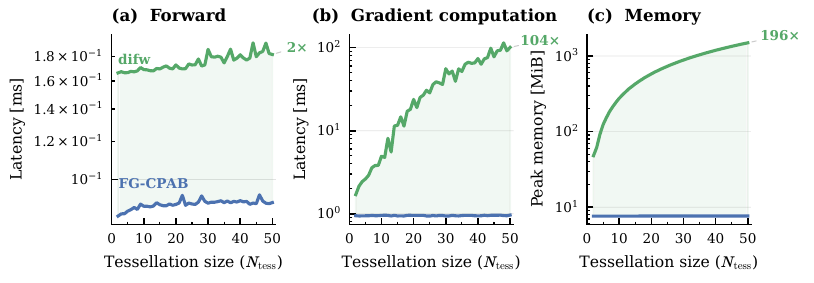}
  \vspace{-0.2cm} 
  \caption{
    \textbf{Performance scaling with tessellation size ($N_\mathrm{tess}$).} 
    Comparison of our optimized 1D CPAB vs.~the DIFW implementation on a single RTX A6000 GPU. 
    \textbf{(a)} Forward pass latency. 
    \textbf{(b)} Gradient computation latency. 
    \textbf{(c)} Peak GPU memory allocation. 
    Our approach yields up to a $2\times$ speedup in the forward pass and a $104\times$ speedup in the gradient computation, while reducing peak memory by $196\times$.
  }
  \label{fig:tess_benchmark_1d_difw}
\end{figure}

\clearpage

\section{Ablations and hyper-parameter analysis}
\label{sec:ablations}

\subsection{Tessellation resolution hyper-parameter analysis}
\label{app:tess_res}

We study the effect of tessellation resolution on CPAB performance, since tessellation size is the main factor controlling the trade-off between expressiveness and computational cost.

\paragraph{Setup.}
We vary the tessellation resolution while keeping the rest of the optimization protocol fixed. We evaluate this both on the IXI brain MRI registration benchmark (\autoref{tab:registratioComparison}, \autoref{fig:ixi_sweep}) and on the synthetic fitting task described in \autoref{app:impl_synthetic_cpab_svf}. The purpose of this ablation is to isolate how much of the final performance comes from access to finer tessellations.

\begin{figure}[htbp]
  \centering
  \includegraphics[width=\linewidth]{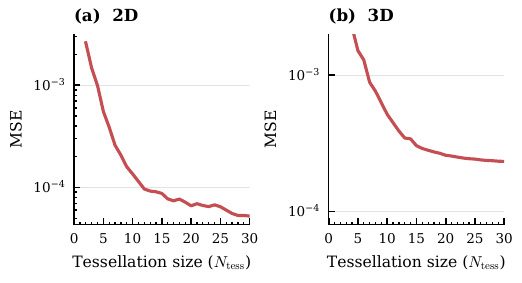}
  \vspace{-0.2cm}
  \caption{
    \textbf{Effect of tessellation resolution on registration performance.}
    Increasing the tessellation resolution improves deformation quality by providing a more expressive piecewise-affine parameterization. This trend is observed across both real and synthetic registration settings, supporting the claim that the scalability unlocked by gradient factorization is important not only for efficiency, but also for final task performance.
  }
  \label{fig:tess_abl}
\end{figure}

\paragraph{Findings.}
\autoref{fig:tess_abl} shows a clear and consistent trend: increasing tessellation resolution improves alignment quality up to the largest resolutions tested. This confirms that finer CPAB parameterizations provide additional local flexibility that is useful for modeling complex deformations. At the same time, the results highlight a central practical limitation of previous CPAB implementations: without an efficient backward pass, these larger tessellations are too slow or memory-intensive to optimize reliably, especially in 3D.

\paragraph{Conclusion.}
The ablation demonstrates that the scalability enabled by gradient factorization is directly tied to model quality. In other words, our method is valuable not only because it makes CPAB faster, but because it enables operation in a more expressive tessellation regime that leads to better registration results.

\subsection{Sparse basis ablation}
\label{app:sparse_abl}

We compare sparse and dense representations of the CPA basis matrix \(\mat{B}\) for 3D transformations on a \(128^3\) grid. As tessellation resolution increases, \(\mat{B}\) grows quickly in both dimensions: the rows correspond to the stacked cell-wise affine coefficients, while the columns correspond to the free CPAB parameters. Since finer tessellations increase both the number of cells and the number of free parameters, a dense basis becomes increasingly expensive, especially in 3D.

\autoref{tab:cpab_sparse_dense_neurips} shows that the sparse basis mainly improves memory usage. At small tessellations, sparse and dense implementations behave similarly, but at tessellation size \(20\), peak memory drops from \(37678\) MiB to \(474\) MiB. Runtime also improves, though more modestly: forward latency decreases from \(85.0\) ms to \(23.5\) ms and backward latency from \(140.3\) ms to \(83.4\) ms.

This trend is expected. The main runtime speedup in our method comes from gradient factorization, which removes parameter-dependent integration from the backward pass. In contrast, sparsifying \(\mat{B}\) mainly reduces the cost of storing and applying the final basis projection. Its primary effect is therefore memory saving, with a secondary but smaller effect on runtime.
\begin{table}[t]
\centering
\small
\setlength{\tabcolsep}{4pt}
\caption{Sparse vs. Dense CPA basis across tessellation size. Forward/backward are mean runtime in ms (lower is better); peak is memory in MiB (lower is better).}
\label{tab:cpab_sparse_dense_neurips}
\begin{tabular}{r|rrr|rrr}
\toprule
& \multicolumn{3}{c|}{Sparse basis} & \multicolumn{3}{c}{Dense basis} \\
\cmidrule(lr){2-4}\cmidrule(l){5-7}
Tess. & Fwd (ms) & Bwd (ms) & Peak (MiB) & Fwd (ms) & Bwd (ms) & Peak (MiB) \\
\midrule
2  & 10.8 & 51.9 &   384 & 10.9 & 51.9 &   384 \\
3  & 11.3 & 54.0 &   384 & 11.3 & 54.1 &   384 \\
4  & 11.5 & 54.9 &   385 & 11.4 & 55.0 &   385 \\
5  & 11.6 & 55.6 &   385 & 11.6 & 55.7 &   385 \\
6  & 11.7 & 56.1 &   386 & 11.7 & 56.3 &   386 \\
7  & 11.9 & 57.0 &   388 & 12.0 & 57.1 &   388 \\
8  & 12.2 & 57.4 &   390 & 12.5 & 57.7 &   390 \\
9  & 12.6 & 58.0 &   392 & 13.0 & 58.4 &   392 \\
10 & 13.8 & 59.6 &   395 & 14.5 & 60.3 &   501 \\
11 & 14.4 & 59.9 &   399 & 15.8 & 61.5 &   914 \\
12 & 15.4 & 62.8 &   403 & 17.8 & 65.1 &  1580 \\
13 & 15.8 & 62.5 &   408 & 19.7 & 65.7 &  2607 \\
14 & 16.9 & 66.2 &   414 & 23.4 & 71.9 &  4140 \\
15 & 17.2 & 65.1 &   423 & 26.9 & 74.1 &  6360 \\
16 & 19.5 & 72.6 &   432 & 34.4 & 86.2 &  9492 \\
17 & 18.4 & 67.6 &   438 & 41.1 & 87.7 & 13818 \\
18 & 20.5 & 75.4 &   450 & 52.0 & 104.8 & 19674 \\
19 & 20.5 & 72.1 &   462 & 64.3 & 113.4 & 27467 \\
20 & 23.5 & 83.4 &   474 & 85.0 & 140.3 & 37678 \\
\bottomrule
\end{tabular}
\end{table}

\subsection{Solver step hyper-parameter analysis}
\label{app:solver_steps}

We study how the number of ODE solver steps affects the trade-off between runtime and registration accuracy. Unless otherwise noted, all results in the paper use 50 CPAB solver steps, matching the standard setting used in prior CPAB implementations.

\paragraph{Setup.}
We use the same synthetic registration protocol as in \autoref{app:impl_synthetic_cpab_svf} and vary only the number of CPAB solver steps. The sweep is performed for 2D transformations with a \(30\times 30\) tessellation, so the goal is to isolate the effect of integration accuracy on optimization behavior without changing the rest of the training setup.

\paragraph{Findings.}
\autoref{tab:transformer-nstepsolver-runtime-accuracy} shows a clear runtime-accuracy trade-off. Very small step counts are fastest per iteration, but they incur substantially worse registration error. Increasing the number of solver steps from 5 to 20 reduces the mean MSE from \(1.42\times 10^{-2}\) to \(1.04\times 10^{-4}\), i.e., by more than two orders of magnitude. Beyond 20 steps, however, the curve largely plateaus. The best mean MSE in the full sweep is \(8.30\times 10^{-5}\) at 65 steps, while 50 steps already achieves \(8.89\times 10^{-5}\). Extending the sweep to 100 steps yields \(8.31\times 10^{-5}\), essentially the same accuracy, but increases mean fitting time from \(0.287\) s to \(0.468\) s and backward latency from \(0.415\) ms to \(0.720\) ms.

\paragraph{Conclusion.}
These results suggest that the paper's default choice of 50 solver steps is a robust operating point: it lies in the accuracy plateau, captures most of the attainable improvement, and avoids the noticeably higher cost of longer integrations.

\begin{table}[htbp]
  \centering
  \scriptsize
  \setlength{\tabcolsep}{5pt}
  \begin{tabular}{@{}rcccc@{}}
    \toprule
    \textbf{$n_{\mathrm{solver}}$} & \textbf{Forward (ms) $\downarrow$} & \textbf{Backward (ms) $\downarrow$} & \textbf{Fit time (s) $\downarrow$} & \textbf{Registration MSE $\downarrow$} \\
    \midrule
    5   & 0.1054 & 0.1822 & 0.2559 & $1.42 \times 10^{-2}$ \\
    10  & 0.1068 & 0.1792 & 0.2490 & $2.27 \times 10^{-3}$ \\
    15  & 0.1080 & 0.2080 & 0.2486 & $3.23 \times 10^{-4}$ \\
    20  & 0.1138 & 0.2279 & 0.2438 & $1.04 \times 10^{-4}$ \\
    25  & 0.1206 & 0.2523 & 0.2482 & $9.44 \times 10^{-5}$ \\
    30  & 0.1248 & 0.2910 & 0.2483 & $9.18 \times 10^{-5}$ \\
    35  & 0.1310 & 0.3201 & 0.2482 & $9.07 \times 10^{-5}$ \\
    40  & 0.1376 & 0.3649 & 0.2558 & $9.72 \times 10^{-5}$ \\
    45  & 0.1441 & 0.3805 & 0.2686 & $8.78 \times 10^{-5}$ \\
    50  & 0.1527 & 0.4145 & 0.2869 & $8.89 \times 10^{-5}$ \\
    55  & 0.1584 & 0.4497 & 0.3066 & $9.00 \times 10^{-5}$ \\
    60  & 0.1659 & 0.4748 & 0.3220 & $8.60 \times 10^{-5}$ \\
    65  & 0.1730 & 0.5147 & 0.3433 & $8.30 \times 10^{-5}$ \\
    70  & 0.1765 & 0.5392 & 0.3599 & $8.47 \times 10^{-5}$ \\
    75  & 0.1838 & 0.5673 & 0.3762 & $8.60 \times 10^{-5}$ \\
    80  & 0.1901 & 0.6109 & 0.4003 & $8.44 \times 10^{-5}$ \\
    85  & 0.1960 & 0.6306 & 0.4134 & $8.47 \times 10^{-5}$ \\
    90  & 0.2011 & 0.6623 & 0.4319 & $8.70 \times 10^{-5}$ \\
    95  & 0.2082 & 0.6966 & 0.4520 & $8.81 \times 10^{-5}$ \\
    100 & 0.2115 & 0.7197 & 0.4678 & $8.31 \times 10^{-5}$ \\
    \bottomrule
  \end{tabular}
  \caption{Full CPAB ODE solver-step sweep over \(n_{\mathrm{solver}}\in\{5,10,\dots,100\}\). Lower values are better. Accuracy improves dramatically from the very low-step regime and then plateaus, while runtime continues to increase with the number of solver steps.}
  \label{tab:transformer-nstepsolver-runtime-accuracy}
\end{table}

\clearpage
\section{Implementation Details}
\label{app:impl_details}

\subsection{Optimization Stability and Practical Expressiveness Compared with Dense SVFs}
\label{app:impl_synthetic_cpab_svf}

\paragraph{Data and deformation generation.}
We follow a simulation-based registration protocol where correspondences are known by construction~\cite{synthmorph}. We evaluate in both 2D and 3D with spatial sizes $512\times512$ and $128\times128\times128$, respectively. For each case, we synthesize a moving image as a sum of 9 random Gaussian blobs (blob standard deviations sampled uniformly from $[0.03, 0.12]$), add Gaussian intensity noise with standard deviation $0.01$, and normalize intensities to $[0,1]$. We then sample a smooth random deformation from a coarse control lattice and upsample it to full resolution (FFD-like deformation model~\cite{nonrigid}), using control sizes $8\times8$ in 2D and $6\times6\times6$ in 3D with deformation amplitude $0.03$. The fixed image is obtained by warping the moving image with this sampled transformation, yielding synthetic moving/fixed pairs with exact ground-truth deformation fields.

\paragraph{Model configuration.}
CPAB uses tessellations of size $15\times15$ in 2D and $15\times15\times15$ in 3D. The dense SVF baseline is parameterized as a full-resolution velocity field and integrated using scaling-and-squaring with 7 squaring steps. This yields a comparison between a compact low-dimensional diffeomorphic parameterization and a standard dense voxel-wise one.

\paragraph{Optimization protocol.}
Both methods are optimized against the same reconstruction objective using Adam. For the fixed-budget comparison reported in the main paper, each method is run for 1000 iterations with 5 restarts per target registration. Per dimension, results are aggregated over 100 independently generated target registrations. This fixed-budget setting is intended to compare practical optimization behavior and time-to-accuracy rather than asymptotic best-case fitting after extensive tuning.

\paragraph{Metrics.}
We report the final mean squared error (MSE) between the recovered image and the target, wall-clock fitting time, and the fraction of sampled spatial locations with non-positive Jacobian determinant after transformation. The Jacobian statistic is included to quantify topology violations.

\begin{figure}[htbp]
  \centering
  \includegraphics[width=\linewidth]{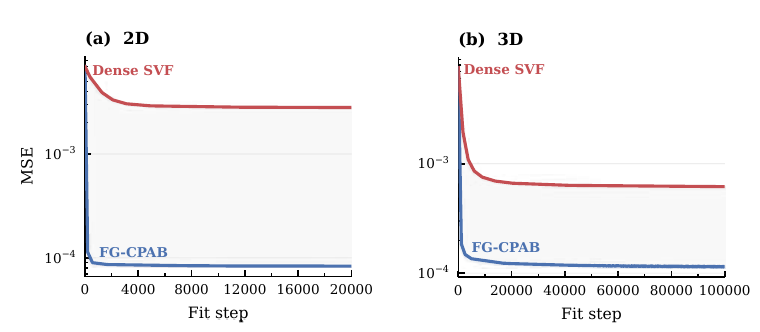}
  \vspace{-0.2cm} 
\caption{
\textbf{Convergence without regularization.}
Optimization trajectories for CPAB and dense SVF on the synthetic fitting task without additional regularization, shown for both 2D and 3D settings. In both cases, CPAB rapidly reaches a low reconstruction error within the first few hundred iterations, whereas dense SVF plateaus early at a substantially higher MSE. This highlights the optimization difficulty of the dense SVF parameterization under a matched iteration budget, despite its much larger number of degrees of freedom.
}
  \label{fig:convergence_noreg}
\end{figure}

\begin{figure}[htbp]
  \centering
  \includegraphics[width=\linewidth]{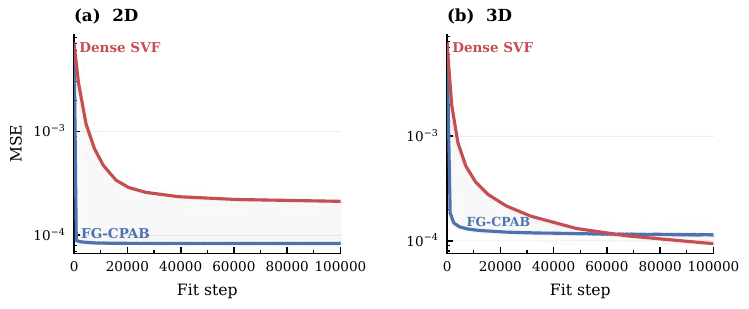}
  \vspace{-0.2cm} 
\caption{\textbf{Convergence with regularization.} Optimization trajectories for CPAB and dense SVF when dense SVF is equipped with standard smoothness regularization, shown for both 2D and 3D settings over an extended budget of 100,000 iterations. CPAB demonstrates extremely rapid initial convergence in both settings, stabilizing within the first few hundred iterations. In 2D (a), CPAB achieves and maintains a consistently lower fitting error. In 3D (b), regularized dense SVF converges much more slowly, but eventually surpasses CPAB to achieve a lower final mean squared error after approximately 50,000 to 60,000 iterations.}
  \label{fig:convergence_reg_svf}
\end{figure}

\paragraph{Fixed-budget behavior.}
Under this matched optimization budget, FG-CPAB consistently achieves lower residual error, shorter fitting times, and fewer topology violations than dense SVF (\autoref{tab:svf_vs_cpab}). One possible explanation is the compact CPAB parameterization, which restricts optimization to the low-dimensional space spanned by $\mat{B}$. In contrast, dense SVF involves a substantially larger number of free parameters. These results suggest that the reduced parameterization may facilitate more efficient optimization while retaining sufficient expressiveness for the evaluated registration tasks.
\paragraph{Extended-budget and regularized dense SVF.}
To separate optimization effects from asymptotic fitting behavior, we also evaluate dense SVF under much larger iteration budgets, both with and without explicit smoothness regularization. \autoref{fig:convergence_noreg} shows that unregularized dense SVF typically plateaus early at a relatively high error, whereas CPAB rapidly reaches a low-error solution. We then equip dense SVF with standard smoothness regularization and optimize for 100{,}000 iterations (\autoref{fig:convergence_reg_svf}). Regularization substantially improves the optimization stability of dense SVF, but convergence remains much slower than for CPAB.

In 2D, regularized dense SVF still does not match CPAB's reconstruction quality within the tested budget. In 3D, regularized dense SVF eventually attains a lower final MSE, but only after approximately $50{,}000$--$60{,}000$ iterations. Since CPAB is also faster per iteration, this crossover occurs only after a very large wall-clock time budget.

\paragraph{Interpretation.}
These experiments indicate that dense SVFs may offer greater asymptotic fitting flexibility in some regimes, but that this flexibility is difficult to realize in practice without strong regularization and a substantially extended optimization budget. CPAB occupies a different and practically attractive point in the trade-off: it provides sufficient diffeomorphic expressiveness for complex deformations while yielding substantially better optimization stability, faster time-to-solution, and fewer topology violations out of the box.

\clearpage

\section{Segmented Checkpointing}
\label{app:segmented_checkpointing}

Segmented checkpointing is the memory mechanism that makes the \(\mathcal{O}(TN)\) adjoint practical on GPU hardware. The reverse sweep requires access to the forward states \(\vect{p}^{(0)},\dots,\vect{p}^{(T)}\), because both the adjoint update and the moment contribution at timestep \(t\) depend on the state and active cell at that timestep. In our implementation these trajectories are thread-private. Storing the full trajectory for every point would therefore require \(\mathcal{O}(T)\) per-thread storage, increase register pressure, and, once registers are exhausted, spill to local memory, reducing occupancy and increasing memory traffic.

\paragraph{Segmented replay strategy.}
Let \(T\) denote the number of integration steps and \(S\) the segment length. We partition the trajectory into
\[
K = \left\lceil \frac{T}{S} \right\rceil
\]
segments, where segment \(k\) covers the timestep range \([kS,\min((k+1)S,T))\). The method then proceeds in two phases.

In the first phase, we integrate forward once and store only the \(K+1\) states at segment boundaries. In the second phase, we process segments in reverse order. For each segment, we restart from its boundary checkpoint, replay the forward dynamics within that segment, and immediately sweep backward through the reconstructed states to propagate the adjoint and accumulate the moments. Once the segment has been processed, its temporary states are discarded.

\paragraph{Memory-compute trade-off.}
The resulting per-thread storage cost is
\[
\mathcal{O}(T/S + S),
\]
consisting of \(\mathcal{O}(T/S)\) checkpoint states and \(\mathcal{O}(S)\) temporary states for the active segment. This replaces the \(\mathcal{O}(T)\) storage required by the full trajectory. The trade-off is a constant-factor recomputation overhead: each state is visited once in the checkpointing pass and once again during replay. Although this does not change the asymptotic complexity, the constant matters in practice because reduced register pressure and lower local-memory traffic can substantially improve GPU occupancy.

\paragraph{Relation to the adjoint factorization.}
Checkpointing changes only how intermediate forward states are materialized; it does not alter the reverse sweep itself. Within each segment, the adjoint recursion and moment accumulation are exactly those of \autoref{eq:adjoint} and \autoref{eq:moment}. Consequently, segmentation changes memory usage and runtime, but not the factorized gradient in \autoref{eq:grad}.

\paragraph{Choice of segment length.}
The segment length \(S\) interpolates between two extremes. When \(S=1\), every timestep is checkpointed: checkpoint count is maximal, but replay buffers are minimal. When \(S\geq T\), the method reduces to a single-segment replay (``no checkpointing'' in \autoref{tab:ckpt_segment_sweep}): checkpoint overhead is minimal, but per-thread replay buffers are largest. The runtime trade-off is therefore typically non-monotonic. Small \(S\) increases checkpoint bookkeeping, whereas large \(S\) increases temporary state, register pressure, and local-memory traffic. Best performance is usually attained at an intermediate \(S\); unless otherwise noted, we use \(S=4\) throughout the paper. \autoref{tab:ckpt_segment_sweep} reports this trade-off empirically.

\begin{table}[htbp]
\centering
\small
\setlength{\tabcolsep}{4pt}
\caption{Checkpoint segment size sweep. Forward/backward are mean runtime in ms (lower is better). Bold indicates the fastest value in each metric.}
\label{tab:ckpt_segment_sweep}
\begin{tabular}{r|rrr}
\toprule
Ckpt seg. & Segments & Fwd (ms) & Bwd (ms) \\
\midrule
2            & 25 & 0.898          & 4.161          \\
3            & 17 & 0.885          & 3.977          \\
4            & 13 & \textbf{0.882} & \textbf{3.860} \\
5            & 10 & 0.885          & 3.905          \\
6            &  9 & 0.891          & 3.899          \\
7            &  8 & 0.889          & 4.302          \\
8            &  7 & 0.893          & 4.644          \\
9            &  6 & 0.889          & 4.699          \\
10           &  5 & 0.887          & 5.056          \\
11           &  5 & 0.891          & 4.905          \\
12           &  5 & 0.887          & 5.073          \\
13           &  4 & 0.887          & 5.246          \\
14           &  4 & 0.885          & 5.198          \\
15           &  4 & 0.892          & 5.284          \\
16           &  4 & 0.885          & 5.683          \\
17           &  3 & 0.893          & 5.976          \\
18           &  3 & 0.905          & 5.868          \\
19           &  3 & 0.901          & 5.828          \\
20           &  3 & 0.898          & 5.923          \\
21           &  3 & 0.897          & 5.956          \\
22           &  3 & 0.897          & 6.080          \\
23           &  3 & 0.900          & 6.231          \\
24           &  3 & 0.898          & 6.468          \\
25           &  2 & 0.897          & 6.773          \\
50 (no ckpt) &  1 & 0.899          & 7.339          \\
\bottomrule
\end{tabular}
\end{table}

\clearpage
\section{Application-Specific Experimental Details}
\label{app:application_details}

We provide additional experimental details for the two
applications considered in the main paper: 3D deformable
registration and neural operator learning. These include
the datasets, model configurations, and optimization protocols.

\subsection{CPAB for 3D deformable registration}
\label{app:impl_ixi_cpab_registration}
\autoref{tab:registratioComparison} is produced using an atlas-to-patient registration protocol on the IXI dataset.

\paragraph{Data and preprocessing.}
We follow the public TransMorph IXI atlas-to-patient protocol~\cite{transmorph}. It contains \(576\) preprocessed T1-weighted MRI volumes, divided into \(403\) training, \(58\) validation, and \(115\) test subjects. Each volume is skull-stripped, affinely aligned, segmented using FreeSurfer, normalized to \([0,1]\), and represented at a resolution of \(160\times192\times224\). Because FG-CPAB requires no offline training, we evaluate it directly on all \(115\) test subjects.

The fixed image is the common atlas distributed with the TransMorph IXI data, originally obtained from the CycleMorph protocol, together with its corresponding segmentation. Registration is optimized from the atlas to each patient using MRI intensities; anatomical labels are used only for evaluation. Dice is evaluated over the \(30\) anatomical labels used by the TransMorph protocol.

\paragraph{Model configuration and optimization.}
In all CPAB configurations, we use a \(20\times20\times20\) tessellation. For the composed variants, multiple CPAB transformations are applied sequentially during each optimization step. The parameters are fitted independently for every test subject using Adam for 200 iterations.

\paragraph{Loss function.}
The objective function comprises a local normalized cross-correlation (LNCC) similarity loss (window size $9$, sample stride $2$, weight $1.0$) and a diffusion regularization penalty (L2 gradient of the displacement field, weight $3.0$).

\subsection{CPAB as a low-dimensional geometry code for operator learning}
\label{app:impl_cpab_operator}

We follow the reaction--diffusion benchmark of DIMON~\cite{dimon}, where solutions on a family of annulus-like domains are transported to a reference circular annulus and a neural operator is conditioned on a low-dimensional geometry code. We keep the DIMON data generation, transport pipeline, and operator architecture unchanged, and vary only the geometry representation.

In the DIMON baseline, geometry is encoded by applying PCA to an LDDMM-based deformation representation. In our variant, we fit a CPAB diffeomorphism from the reference domain to each target domain and use the CPAB parameters directly as the geometry code. We use a \(6 \times 6\) CPAB tessellation, yielding \(122\) parameters. For a fair comparison, we also set the DIMON baseline to \(122\) PCA coefficients. Hence both methods use geometry codes of identical dimension, and differences in performance reflect the quality of the underlying geometry parameterization rather than code size.

\newpage

\section{Nodal Sparse Basis}
\label{app:nodal_basis}

The basis \(\mat{B}\) introduced in \autoref{sec:background} can be constructed efficiently through a nodal parameterization of the CPA velocity field. Instead of working directly with cell-wise affine coefficients, we first represent the velocity field by its values at the tessellation vertices and then impose the CPAB constraints in this nodal space. This yields a sparse construction of a basis spanning the same space as the constraint-based construction.

\paragraph{Nodal parameterization.}
Let \(n_V\) denote the number of vertices in the tessellation, and let
\(\vect{u}\in\mathbb{R}^{K n_V}\) be the stacked vector of nodal velocities, with one \(K\)-dimensional velocity vector per vertex. Let \(\vect{a}\in\mathbb{R}^{D}\) denote the stacked vector of all cell-wise affine coefficients, where \(D\) is the total number of affine coefficients across all cells. Since the velocity field is affine inside each simplex, the affine coefficients on a cell are uniquely determined by the velocities at that cell's \(K+1\) vertices. Therefore, the global affine parameter vector can be written as
\begin{align}
\vect{a} = \mat{N}\vect{u},
\qquad
\mat{N}\in\mathbb{R}^{D\times K n_V},
\label{eq:nodal_map}
\end{align}
where \(\mat{N}\) is a sparse linear map from nodal velocities to cell-wise affine parameters.

\paragraph{Construction of the nodal map.}
The matrix \(\mat{N}\) is assembled cell by cell from the mesh geometry. For a simplex \(c\), let
\[
\widetilde{\vect{x}}_{c,0},\dots,\widetilde{\vect{x}}_{c,K}\in\mathbb{R}^{K+1}
\]
denote its vertices in homogeneous coordinates, and let
\[
\vect{v}_{c,0},\dots,\vect{v}_{c,K}\in\mathbb{R}^{K}
\]
be the corresponding nodal velocities. Inside the simplex, the CPA velocity field is represented as
\[
\vect{v}_c(\vect{x}) = \mat{A}_c \widetilde{\vect{x}},
\qquad
\mat{A}_c\in\mathbb{R}^{K\times(K+1)}.
\]
The local affine matrix \(\mat{A}_c\) is determined by the interpolation conditions
\[
\mat{A}_c \widetilde{\vect{x}}_{c,j} = \vect{v}_{c,j},
\qquad j=0,\dots,K.
\]
Collecting the homogeneous vertex coordinates and nodal velocities into
\[
\mat{X}_c =
\begin{bmatrix}
\widetilde{\vect{x}}_{c,0} & \cdots & \widetilde{\vect{x}}_{c,K}
\end{bmatrix}
\in\mathbb{R}^{(K+1)\times(K+1)},
\qquad
\mat{V}_c =
\begin{bmatrix}
\vect{v}_{c,0} & \cdots & \vect{v}_{c,K}
\end{bmatrix}
\in\mathbb{R}^{K\times(K+1)},
\]
gives
\[
\mat{A}_c \mat{X}_c = \mat{V}_c.
\]
Since the simplex vertices are affinely independent, \(\mat{X}_c\) is invertible, and thus
\[
\mat{A}_c = \mat{V}_c \mat{X}_c^{-1}.
\]

This is the local interpolation rule in matrix form. To express it as a linear operator on stacked vectors, we apply the standard identity
\[
\operatorname{vec}(\mat{U}\mat{M}) = (\mat{M}^{\top}\otimes \mat{I})\operatorname{vec}(\mat{U}),
\]
which yields
\[
\operatorname{vec}(\mat{A}_c)
=
(\mat{X}_c^{-T}\otimes \mat{I}_K)\operatorname{vec}(\mat{V}_c).
\]
Hence the local nodal-to-affine operator is
\[
\mat{N}_c = \mat{X}_c^{-T}\otimes \mat{I}_K.
\]
The global sparse matrix \(\mat{N}\) is obtained by assembling these local operators over all cells according to the mesh connectivity. Its sparsity follows from locality: each cell depends only on its own \(K+1\) vertices, so each row of \(\mat{N}\) contains nonzero entries only in the columns corresponding to those local nodal degrees of freedom.

\paragraph{Constraint enforcement in nodal coordinates.}
In the standard CPAB formulation, valid affine coefficients must satisfy the global linear constraint system
\[
\mat{L}\vect{a} = \vect{0},
\]
which enforces continuity across shared cell boundaries together with any imposed boundary conditions. Substituting the nodal parameterization from \autoref{eq:nodal_map} gives
\begin{align}
\mat{L}\mat{N}\vect{u} = \vect{0}.
\label{eq:nodal_constraint}
\end{align}
Defining
\begin{align}
\mat{C} \coloneqq \mat{L}\mat{N},
\end{align}
the admissible nodal configurations are exactly those in the null space of \(\mat{C}\). Let
\begin{align}
\mat{Z}\in\mathbb{R}^{K n_V\times d}
\end{align}
be a basis for \(\operatorname{null}(\mat{C})\), so that every valid nodal vector can be written as
\begin{align}
\vect{u} = \mat{Z}\vect{\theta},
\qquad
\vect{\theta}\in\mathbb{R}^d.
\label{eq:nodal_null}
\end{align}

\paragraph{Resulting sparse CPA basis.}
Combining \autoref{eq:nodal_map} and \autoref{eq:nodal_null} yields
\begin{align}
\vect{a} = \mat{N}\mat{Z}\vect{\theta}.
\label{eq:nodal_basis_product}
\end{align}
Hence the CPA basis can be written as
\begin{align}
\mat{B} = \mat{N}\mat{Z}.
\label{eq:nodal_basis}
\end{align}
This shows that the low-dimensional parameter vector \(\vect{\theta}\) can be interpreted as coordinates in a constrained nodal basis, while \(\mat{B}\) maps these coordinates to the cell-wise affine representation used throughout the paper.

The main advantage of this construction is sparsity: \(\mat{N}\) is local by design, and the resulting basis inherits the mesh structure of the tessellation. Crucially, on a standard simplex mesh, the nodal formulation automatically satisfies continuity across shared cell boundaries. Consequently, the vast majority of constraints in \(\mat{C} = \mat{L}\mat{N}\) evaluate exactly to zero. The only remaining active constraints are global boundary conditions (e.g., zero velocity at the domain boundaries). Finding the null space \(\mat{Z}\) of this drastically reduced system geometrically amounts to identifying the unconstrained interior nodes, yielding a natural sparse basis matrix. In practice, simply thresholding numerical noise from standard dense linear algebra routines fully recovers this exact physical sparsity. This provides an efficient way to assemble \(\mat{B}\) as a sparse matrix while exactly preserving the same globally continuous CPA velocity fields as the direct algebraic construction.

\paragraph{Remark.}
The null-space basis \(\mat{Z}\) can often be constructed
directly from the nodal constraints, avoiding an SVD.
For more general constraints, one can remove the redundant
rows of \(\mat{C}\) before computing its null space numerically,
substantially reducing the cost of the SVD.
Although sparsity is not guaranteed in general, the resulting
basis can remain sparse in practice for certain constraint
structures, enabling efficient storage and computation.
\newpage

\section{Proof of the Telescoping Property of the Discrete Adjoint}
\label{app:telescoping}

We prove the claim made in \autoref{subsec:moments} that summing the sensitivity contraction identity~\autoref{eq:explicit_adjoint_sensitivity} over all timesteps produces a telescoping series in which all propagated terms cancel exactly, leaving only the accumulated direct contributions.

For notational brevity, fix a single point \(\vect{p}_n\) (the argument is identical and independent for every point) and a single parameter component \(\theta_i\). Define the scalar quantities
\[
s^{(t)} \coloneqq \vect{\lambda}^{(t)\trans} \pd{\vect{p}^{(t)}}{\theta_i},
\qquad
d^{(t)} \coloneqq \vect{\lambda}^{(t+1)\trans} \Bigl( h\,\mat{B}_{i,c}\,\widetilde{\vect{p}}_{\mathrm{mid}}^{(t)}
+ \tfrac{h^2}{2}\,\mat{A}_{c,\mathrm{lin}}\,\mat{B}_{i,c}\,\widetilde{\vect{p}}^{(t)} \Bigr),
\]
where all quantities on the right-hand side of \(d^{(t)}\) are evaluated at the appropriate cell \(c\) that the point occupies at timestep \(t\).

The identity~\autoref{eq:explicit_adjoint_sensitivity} then reads
\begin{align}\label{eq:contracted_app}
  s^{(t+1)} = d^{(t)} + s^{(t)}
  \qquad\text{for each timestep } t = 0,1,\dots,T-1.
\end{align}

Summing~\autoref{eq:contracted_app} from \(t=0\) to \(t=T-1\) yields
\[
\sum_{t=0}^{T-1} s^{(t+1)} = \sum_{t=0}^{T-1} d^{(t)} + \sum_{t=0}^{T-1} s^{(t)}.
\]

The left-hand side expands to
\[
s^{(1)} + s^{(2)} + \cdots + s^{(T)},
\]
while the second sum on the right-hand side expands to
\[
s^{(0)} + s^{(1)} + \cdots + s^{(T-1)}.
\]
Rearranging immediately gives the telescoping cancellation
\begin{align}\label{eq:telescoping}
  s^{(T)} - s^{(0)} = \sum_{t=0}^{T-1} d^{(t)}.
\end{align}

Two boundary conditions complete the proof:

1. The initial position \(\vect{p}^{(0)} = \vect{p}_0\) is the fixed input point and therefore independent of the parameters \(\vect{\theta}\). Consequently,
   \[
   \pd{\vect{p}^{(0)}}{\theta_i} = \vect{0} \quad\implies\quad s^{(0)} = 0.
   \]

2. At the final timestep the adjoint is defined by the loss gradient,
   \[
   \vect{\lambda}^{(T)} = \pd{\mathcal{L}}{\vect{p}^{(T)}},
   \]
   so
   \[
   s^{(T)} = \vect{\lambda}^{(T)\trans} \pd{\vect{p}^{(T)}}{\theta_i} = \pd{\mathcal{L}}{\theta_i}.
   \]

Substituting these two facts into~\autoref{eq:telescoping} produces
\[
\pd{\mathcal{L}}{\theta_i} = \sum_{t=0}^{T-1} d^{(t)}.
\]

The right-hand side is precisely the sum of all direct (local) contributions of \(\theta_i\) across the trajectory. These contributions are exactly the terms that, after applying the Frobenius inner-product identity and summing over all points \(n\) that visit cell \(c\) at timestep \(t\), assemble into the moment matrix \(\mat{M}_c\) of~\autoref{eq:moment}. The final gradient is then recovered by the cheap projection~\autoref{eq:grad}.

Because the derivation relies only on the chain rule and the definition of the exact discrete adjoint of the midpoint integrator, the cancellation is algebraic and holds to machine precision (no discretization error is introduced).

\section{Why Translation Does Not Enter the State Jacobian}
\label{app:jacobian_translation}

For an arbitrary spatial dimension $K$, consider a fixed cell $c$
and decompose the affine velocity matrix as
\[
\mat{A}_c = [\mat{L}_c\;\;\vect{b}_c],
\qquad
\mat{L}_c\in\mathbb{R}^{K\times K},\quad
\vect{b}_c\in\mathbb{R}^{K}.
\]
For homogeneous coordinates
$\widetilde{\vect{p}}=(\vect{p}^{\top},1)^{\top}$,
the velocity field is
\[
\mat{A}_c\widetilde{\vect{p}}
= \mat{L}_c\vect{p} + \vect{b}_c.
\]

Starting from the midpoint integrator in~\autoref{eq:midpoint},
\[
\vect{p}_{\mathrm{mid}}
= \Bigl(\mat{I} + \frac{h}{2}\mat{L}_c\Bigr)
  \vect{p}^{(t)}
  + \frac{h}{2}\vect{b}_c,
\]
and therefore
\[
\vect{p}^{(t+1)}
= \Bigl(\mat{I} + h\mat{L}_c
  + \frac{h^2}{2}\mat{L}_c^2\Bigr)\vect{p}^{(t)}
  + \Bigl(h\mat{I} + \frac{h^2}{2}\mat{L}_c\Bigr)\vect{b}_c.
\]

Hence, the state Jacobian of one midpoint step is
\[
\pd{\vect{p}^{(t+1)}}{\vect{p}^{(t)}}
= \mat{I} + h\mat{L}_c + \frac{h^2}{2}\mat{L}_c^2,
\]
where $\mat{I}\in\mathbb{R}^{K\times K}$.
Thus, the Jacobian depends only on the linear block
$\mat{L}_c=\mat{A}_{c,\mathrm{lin}}$,
regardless of the spatial dimension.
The translation $\vect{b}_c$ contributes only additive
terms independent of $\vect{p}^{(t)}$, whose derivatives
\emph{w.r.t.}~$\vect{p}^{(t)}$ vanish.
\newpage

\section{Integrator mismatch}
\label{app:integrator_mismatch}

\paragraph{Source of the mismatch and relation to prior CPAB implementations.}
As discussed in \autoref{sec:method}, our forward pass uses the matrix-exponential update, whereas the gradient computation uses the discrete adjoint of the explicit midpoint scheme in \autoref{eq:midpoint}. Consequently, the resulting gradient is the exact derivative of the midpoint-discretized dynamics, but only an approximation to the derivative of the matrix-exponential forward computation.

This numerical pairing is not introduced by FG-CPAB. Reference CPAB implementations similarly combine a matrix-exponential forward transformation with a midpoint-discretized sensitivity computation~\cite{cpab2017,libcpab}. We retain the same forward and gradient discretizations to ensure a fair comparison with prior CPAB methods. Thus, the comparison evaluates our factorized organization of the gradient computation without introducing a different integration scheme as an additional source of performance or accuracy differences.

\paragraph{Effect of inconsistent cell assignments.}
We first consider the discrepancy caused by using different cells for the same state. Let \(\widetilde{\mathbf A}_1\) and \(\widetilde{\mathbf A}_2\) denote the augmented matrices representing the velocity fields in two neighboring cells, and consider
\begin{align}
\widetilde{\vect{p}}_1^{(t+1)}
&=
\exp(h\widetilde{\mat{A}}_1)\widetilde{\vect{p}}^{(t)},
&
\widetilde{\vect{p}}_2^{(t+1)}
&=
\exp(h\widetilde{\mat{A}}_2)\widetilde{\vect{p}}^{(t)}.
\end{align}
A Taylor expansion gives
\begin{align}
\widetilde{\vect{p}}_1^{(t+1)}
-
\widetilde{\vect{p}}_2^{(t+1)}
=
h\bigl(\widetilde{\mat{A}}_1-\widetilde{\mat{A}}_2\bigr)
\widetilde{\vect{p}}^{(t)}
+
\mathcal{O}(h^2).
\end{align}
Therefore, for fixed cell matrices and a bounded state (i.e., $\|\widetilde{p}^{(t)}\|$ remains bounded as $h \to 0$), an inconsistent cell choice can introduce an \(O(h)\) local discrepancy.

\paragraph{Effect of different integrators within a fixed cell.}
We next compare the matrix-exponential and midpoint updates when both use the same cell matrix \(\widetilde{\mat{A}}\). The midpoint update can be written as
\begin{align}
\widetilde{\vect{p}}_{\mathrm{mid}}^{(t+1)}
=
\left(
\mat{I}
+
h\widetilde{\mat{A}}
+
\tfrac{h^2}{2}\widetilde{\mat{A}}^2
\right)
\widetilde{\vect{p}}^{(t)}.
\end{align}
Meanwhile,
\begin{align}
\exp(h\widetilde{\mat{A}})
=
\mat{I}
+
h\widetilde{\mat{A}}
+
\tfrac{h^2}{2}\widetilde{\mat{A}}^2
+
\mathcal{O}\!\left(h^3
\lVert\widetilde{\mat{A}}\rVert^3\right).
\end{align}
Hence,
\begin{align}
\widetilde{\vect{p}}_{\exp}^{(t+1)}
-
\widetilde{\vect{p}}_{\mathrm{mid}}^{(t+1)}
=
\mathcal{O}\!\left(
h^3\lVert\widetilde{\mat{A}}\rVert^3
\right).
\end{align}

For a fixed cell matrix over a finite integration interval,
the \(\mathcal{O}(h^3)\) local error yields an
\(\mathcal{O}(h^2)\) global error.
Thus, the local integrator mismatch is of higher order
than the potential \(\mathcal{O}(h)\) discrepancy caused
by inconsistent cell assignments.

\paragraph{Empirical gradient validation.}
The remaining question is whether this approximation is sufficiently accurate for gradient-based optimization. We therefore compared the gradient produced by our backward pass with a central finite-difference estimate of the gradient of the matrix-exponential forward computation. We measured cosine similarity and relative error across multiple tessellation configurations.

For the standard setting used throughout the paper, with \(50\) solver steps, the cosine similarity was above \(0.998\), while the relative error was below \(0.05\). The approximate gradient is therefore closely aligned with the finite-difference reference gradient in the regime used in our experiments.

\paragraph{Conclusion.}
The forward and backward discretizations are not algebraically identical, and our gradient should accordingly be interpreted as an approximation to the derivative of the matrix-exponential forward computation. However, the mismatch is higher order within a fixed cell, follows the numerical treatment used by prior CPAB implementations, and produces gradients that are strongly aligned with finite-difference estimates. Empirically, it does not hinder stable first-order optimization in any of our experiments.
\clearpage

\section{Why Gradient-Based Optimization Is Needed: Comparison with MCMC}
\label{app:mcmc_comparison}

The original CPAB work considered MCMC-based optimization (Metropolis algorithm with a Gaussian proposal) for standalone registration problems, while its journal extension subsequently introduced gradient-based optimization~\cite{cpab,cpab2017}. Although MCMC can be effective in low-dimensional settings, it is not well suited to the fine-tessellation applications targeted in this work for two main reasons.

\paragraph{End-to-end differentiability.}
CPAB was originally introduced as a standalone transformation model outside modern automatic-differentiation frameworks. In many recent applications, however, a CPAB transformation is used as a differentiable component within a larger model. Training such a model requires propagating loss gradients through the CPAB parameters to preceding network components. MCMC performs inference over the CPAB parameters but does not provide the vector--Jacobian products required for end-to-end backpropagation.

\paragraph{Scaling with the parameter dimension.}
Standard MCMC methods, such as Metropolis sampling with a Gaussian proposal, generally become increasingly difficult to apply as the parameter dimension grows. Early CPAB experiments using MCMC involved relatively small parameter spaces; for example, the landmark experiment in the original CPAB work used \(d=38\) parameters. In contrast, our synthetic benchmarks use \(842\) parameters in 2D and \(8{,}232\) parameters in 3D, while the IXI configuration uses \(20{,}577\) parameters per CPAB transformation. This fine-tessellation regime strongly favors gradient-based optimization.

\paragraph{Empirical comparison.}
To verify this empirically, we benchmarked the official CPAB MCMC implementation using the same IXI atlas-to-patient registration setup as FG-CPAB. We allocated MCMC a wall-clock optimization budget \(100\times\) larger than that used by FG-CPAB. Nevertheless, MCMC achieved a Dice score of only \(0.453\), compared with \(0.760\) for gradient-based FG-CPAB.

These results do not imply that MCMC is unsuitable for every CPAB application. Rather, they show that it is uncompetitive for the high-dimensional, fine-tessellation problems considered here and cannot replace gradient computation when CPAB is used within an end-to-end differentiable model.

\clearpage

\section{Assets, Licenses, and Terms of Use}
\label{app:licenses}

This appendix is provided to document the external assets used in this work, including datasets, codebases, and benchmark implementations. 

\subsection{Public datasets and benchmark data}

\begin{table}[htbp]
\centering
\scriptsize
\setlength{\tabcolsep}{4pt}
\caption{Datasets and benchmark data used in the paper.}
\label{tab:asset_datasets}
\resizebox{\linewidth}{!}{%
\begin{tabular}{p{2.2cm}p{2.5cm}p{2.4cm}p{2.0cm}p{2.4cm}}
\toprule
\textbf{Asset} & \textbf{How used in this paper} & \textbf{Citation} & \textbf{Version} & \textbf{License} \\
\midrule
IXI brain MRI dataset &
3D deformable registration benchmark &
\cite{ixi_dataset} &
\href{https://brain-development.org/ixi-dataset/}{link} 26.4.26 &
Creative Commons CC BY-SA 3.0 license \\

DIMON reaction--diffusion benchmark data / pipeline &
Operator-learning experiment in \autoref{subsec:cpab_operator} &
\cite{dimon} &
Commit - \href{https://github.com/MinglangYin/DIMON/tree/a22ed33f25939603f8605914e39be7de082c1b1b}{a22ed33} &
MIT License \\

\bottomrule
\end{tabular}%
}
\end{table}

\subsection{External code, software, and baseline implementations}

\begin{table}[htbp]
\centering
\scriptsize
\setlength{\tabcolsep}{4pt}
\caption{External codebases, libraries, and baseline implementations used by the project.}
\label{tab:asset_software}
\resizebox{\linewidth}{!}{%
\begin{tabular}{p{2.4cm}p{2.6cm}p{2.3cm}p{2.1cm}p{2.2cm}}
\toprule
\textbf{Asset} & \textbf{How used in this paper} & \textbf{Citation} & \textbf{Version / commit} & \textbf{License} \\
\midrule
\texttt{libcpab} &
Primary CPAB baseline and reference implementation &
\cite{libcpab} &
Commit - \href{https://github.com/freifeld/cpabDiffeo/tree/22df6cdbd7111b9ae3e7f1c0e31ff85e92d281a6}{22df6cd} &
 MIT license \\

\texttt{DIFW} &
1D performance baseline in \autoref{sec:appendix_performance} &
\cite{martinez2022closed} &
Commit - \href{https://github.com/imartinezl/difw/tree/f970dd3049c4099c2753871e26d66f753e3a07c0}{f970dd3} &
MIT license \\

VoxelMorph &
Dense SVF baseline / runtime comparison &
\cite{voxelmorph} &
Commit - \href{https://github.com/voxelmorph/voxelmorph/tree/6fea3e8722c3db7edeffc663700acb0d5bcbb3c3}{6fea3e8} &
Apache-2.0 license \\

\texttt{torch-lddmm} &
LDDMM runtime baseline &
\cite{lddmm} &
Commit - \href{https://github.com/brianlee324/torch-lddmm/tree/86d5c43f779563a26356c0b2eb84e85c7b98a1ed}{86d5c43} &
 Apache-2.0 license \\

\bottomrule
\end{tabular}%
}
\end{table}
\end{document}